%% file: main.tex
\documentclass[sigconf,10pt]{acmart}

\acmYear{2023}\copyrightyear{2023}
\setcopyright{acmcopyright}
\acmConference[MobiCom '23]{The 29th Annual International Conference On Mobile Computing And Networking}{2-6 Oct 2023}{Madrid, Spain}
\acmBooktitle{The 29th Annual International Conference On Mobile Computing And Networking (MobiCom '23), 2-6 Oct 2023, Madrid, Spain}

\acmPrice{}
\usepackage{times}
\usepackage{soul}
\usepackage{url}
\usepackage[utf8]{inputenc}
\usepackage{caption}
\usepackage{tikz}
\usepackage{amsmath,amsfonts, bm}
\usepackage{algorithm}
\usepackage{algorithmicx}
\usepackage{algpseudocode}  % 这个和algorithmic不兼容，用了就要报错，好多莫名其妙的错误！！！！！
\usepackage{graphicx}
\usepackage{textcomp}
\usepackage{booktabs} % For formal tables
\usepackage{siunitx}
\usepackage{xspace,url}

\usepackage{color}
\usepackage{colortbl}
\usepackage{multirow}
\usepackage{multicol}
\usepackage{makecell}
\usepackage{tabularx}
\usepackage{listings}
\usepackage{graphicx}
\usepackage{caption}
\usepackage{enumitem}
\usepackage{adjustbox}
\usepackage{subcaption}

\usepackage{filecontents}
\usepackage[colorinlistoftodos]{todonotes}

\definecolor{airforceblue}{rgb}{0.36, 0.54, 0.66}
\newcommand{\name}{AdaptiveNet\xspace} % name of the tool

\newcommand{\ie}{\textit{i}.\textit{e}.~}
\newcommand{\eg}{\textit{e}.\textit{g}.~}

\author{Hao Wen}
\affiliation{Institute for AI Industry Research (AIR), Tsinghua University}

\author{Yuanchun Li}\authornote{Yuanchun Li (liyuanchun@air.tsinghua.edu.cn) is the corresponding author. Yuanchun Li and Yunxin Liu are also affiliated with Shanghai AI Laboratory.}
\affiliation{Institute for AI Industry Research (AIR), Tsinghua University}
\author{Zunshuai Zhang}
\affiliation{Shanghai University}

\author{Shiqi Jiang}
\affiliation{Microsoft Research}

\author{Xiaozhou Ye}
\affiliation{AsiaInfo Technologies (China), Inc.}

\author{Ye Ouyang}
\affiliation{AsiaInfo Technologies (China), Inc.}

\author{Ya-Qin Zhang}
\affiliation{Institute for AI Industry Research (AIR), Tsinghua University}

\author{Yunxin Liu}
\affiliation{Institute for AI Industry Research (AIR), Tsinghua University}
\renewcommand{\shortauthors}{Wen et al.}

\begin{document}
\title{\name: Post-deployment Neural Architecture Adaptation for Diverse Edge Environments}

% \title{\name: Scaling Deep Learning Model to Diverse Edge Devices via Post-deployment Neural Architecture Adaptation}

% \title{\name: Adapting Pretrained Model to Diverse Edge Environments via On-device Neural Architecture Search}

%    <concept>
%        <concept_id>10010520.10010553.10010562</concept_id>
%        <concept_desc>Computer systems organization~Embedded systems</concept_desc>
%        <concept_significance>500</concept_significance>
%        </concept>
%    <concept>
%        <concept_id>10010147.10010257.10010293.10010294</concept_id>
%        <concept_desc>Computing methodologies~Neural networks</concept_desc>
%        <concept_significance>500</concept_significance>
%        </concept>
%  </ccs2012>

% \ccsdesc[500]{Computing methodologies~Neural networks}

% \keywords{Edge Device, Deep Learning, Neural Architecture Adaptation, Model Scaling, Model Adaptation}

\begin{abstract}
Deep learning models are increasingly deployed to edge devices for real-time applications. To ensure stable service quality across diverse edge environments, it is highly desirable to generate tailored model architectures for different conditions. However, conventional pre-deployment model generation approaches are not satisfactory due to the difficulty of handling the diversity of edge environments and the demand for edge information. In this paper, we propose to adapt the model architecture after deployment in the target environment, where the model quality can be precisely measured and private edge data can be retained. To achieve efficient and effective edge model generation, we introduce a pretraining-assisted on-cloud model elastification method and an edge-friendly on-device architecture search method. Model elastification generates a high-quality search space of model architectures with the guidance of a developer-specified oracle model. Each subnet in the space is a valid model with different environment affinity, and each device efficiently finds and maintains the most suitable subnet based on a series of edge-tailored optimizations. Extensive experiments on various edge devices demonstrate that our approach is able to achieve significantly better accuracy-latency tradeoffs (e.g. 46.74\% higher on average accuracy with a 60\% latency budget) than strong baselines with minimal overhead (13 GPU hours in the cloud and 2 minutes on the edge server).
\end{abstract}

\maketitle

\input{tex/intro}

\input{tex/background}

% \input{tex/measurements}
\input{tex/approach}

\input{tex/implementation}
\input{tex/experiment}
\input{tex/conclusion}

% \bibliographystyle{plain}
% \bibliography{reference}
\bibliographystyle{ACM-Reference-Format}
\bibliography{reference}

\end{document}

%% file: tex/intro.tex
\section{Introduction}

% \yuanchun{abstract and introduction are ready for suggestions.}
% \renewcommand{\shortauthors}{Wen et al.}
% Edge DNN is popular
Deep learning has enabled and enhanced many intelligent applications at the edge, such as driving assistance \cite{VIeye,mobiad}, face authentication \cite{face, face_infocom}, video surveillance \cite{loki,deepcache}, speech recognition \cite{google_speech,nipsspeech}, etc.
% Among the applications, many are real-time tasks that require to get response immediately.
Due to latency and privacy considerations, it is an increasingly common practice to deploy the models to edge devices \cite{pecam, mistify}, so that the models can be invoked directly without transmitting data to the server.
% With the help of model compression techniques \cite{}, the computation and size of DNN models can be greatly reduced to fit into resource-constrained edge environments.

% The challenge of diversity
% Environment diversity is a main challenge of deploying DNNs to the edge.
The diversity of execution environments is a unique characteristic of edge devices as compared with the cloud.
For example, a video app may deploy a super-resolution model on millions of smartphones, ranging from low-end devices to high-end ones, and their computational capacity may differ by up to 20 times. Generating a model for each type of device to guarantee the user experience is very time-consuming; an object detection model may run on different kinds of driving assistance systems, and the computational power may range from 20 to 1000 TOPS. To guarantee safety, the model inference usually has a strict latency budget.
%For example, an object detection model may run on different smart cameras and edge servers, and an AI-enhanced online conferencing app may be accessed on different desktops and mobile devices.
Even on the same type of devices, the model execution environments may also vary across instances and change over time due to different hardware states and concurrent processes.
To provide a good and uniform user experience, developers are usually required to generate tailored models for diverse edge environments.

% % common practice: model compression
% The current practice to handle environment diversity is to generate different model variants using model compression techniques, and deploy the appropriate version to each edge environment.
% The drawback of such a solution is two-fold.
% First, the limited number of model variants is difficult to handle the large range of edge environment diversity. \xx
% Second, generating model variants introduces extra cost of training and management. \xx

% limitation of existing methods
There are many techniques proposed to automatically generate tailored models according to the target environments.
Most of them are cloud-based approaches, in which the models are determined on the cloud side before distributing to edge devices. We call them \textbf{``pre-deployment approaches''} in this paper.
Neural Architecture Search (NAS) \cite{proxylessnas,ofa,darts-gradient-based,nasrl,nas-rl,evolutionarysearch} is the most popular technique of this type due to its superior flexibility to change network architectures.
It typically requires collecting information (about computational resources, runtime conditions, data distribution, etc.) from the target environments to guide the model architecture search and training processes in the cloud.

\begin{figure*}
    \centering
    \begin{subfigure}[b]{0.48\textwidth}
        \centering
        \includegraphics[width=8cm]{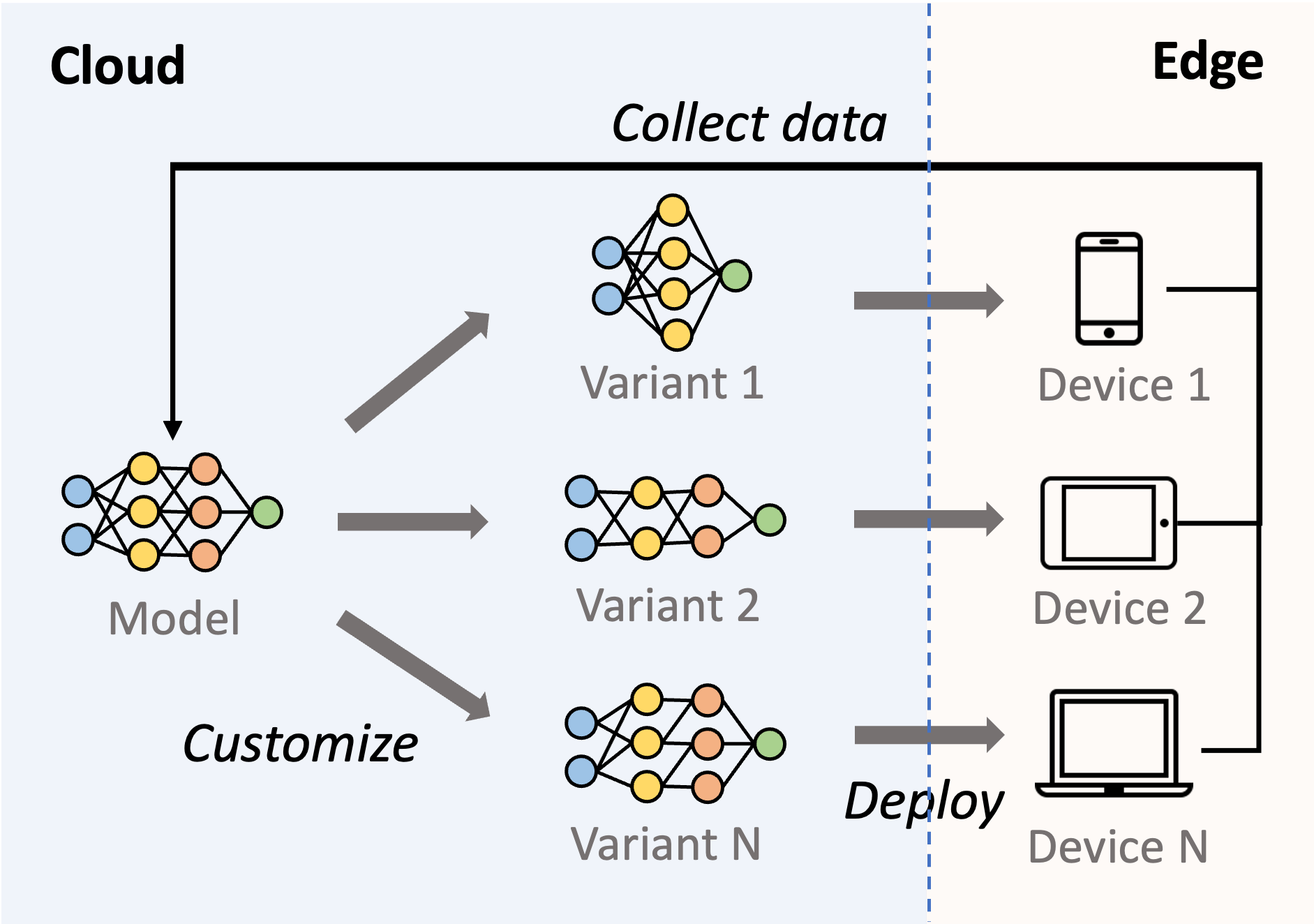}
        \caption{Pre-deployment on-cloud model generation (conventional).}
        \label{fig:process_comparison:cloud}
    \end{subfigure}
    ~~
    \begin{subfigure}[b]{0.48\textwidth}
        \centering
        \includegraphics[width=8cm]{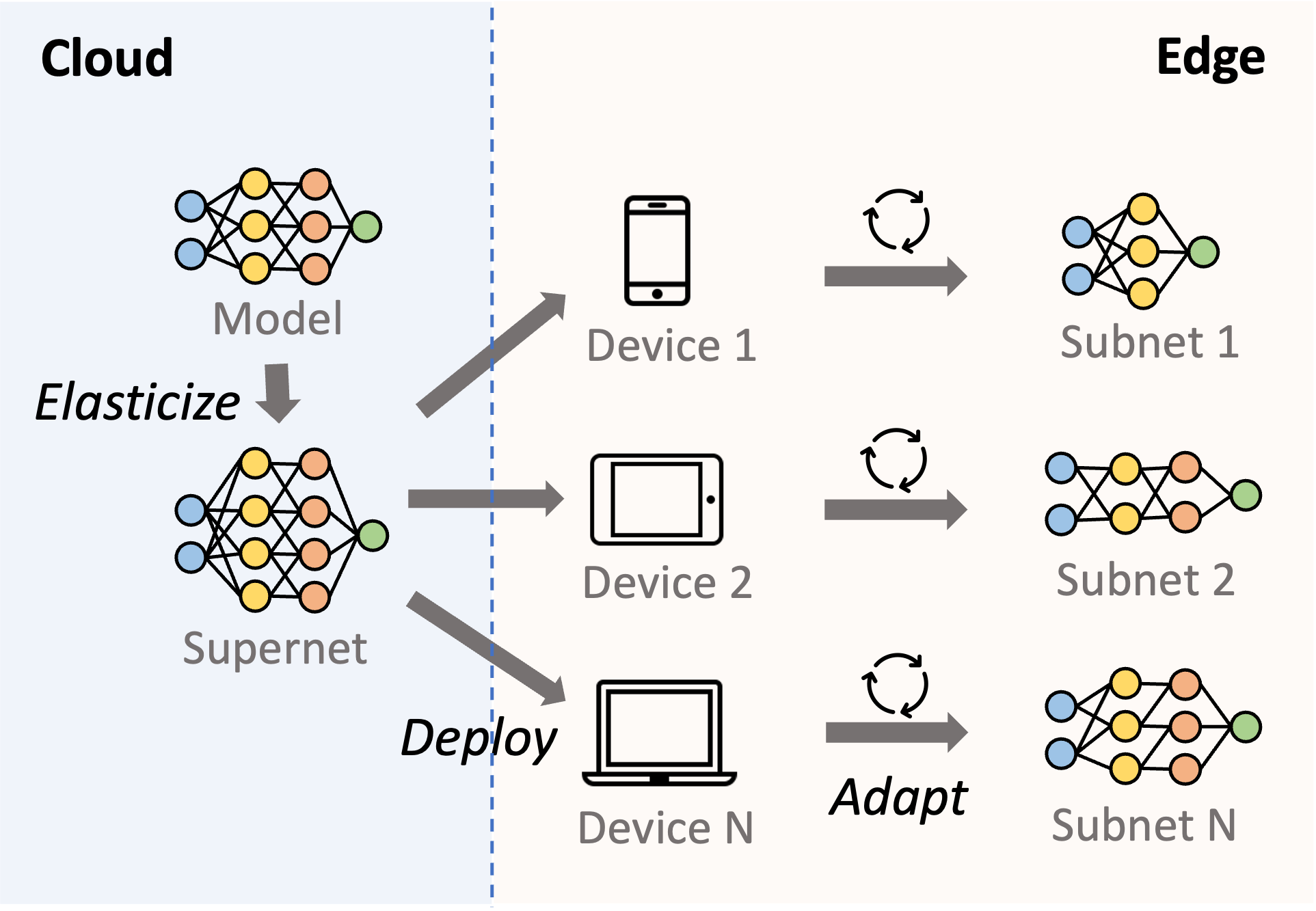}
        \caption{Post-deployment on-device model adaptation (ours).}
        \label{fig:process_comparison:edge}
    \end{subfigure}
    \vspace{-0.3cm}
    \caption{Comparison of pre-deployment and post-deployment model generation approaches.}
    \label{fig:process_comparison}
\end{figure*}

Despite the effectiveness to find optimal model architecture based on the target environment, NAS approaches are less practical in many edge/mobile scenarios where the model execution environments may be very diverse and dynamic.
Searching and maintaining the optimal model architecture in the cloud for each edge would be very compute- and labor-intensive.
Thus, \emph{a more economic and ideal solution is to let the model self-adapt to the target environment after deployment,} which we call \textbf{``post-deployment approach''} to distinguish with the conventional methods, as illustrated in Figure~\ref{fig:process_comparison}.
Doing so brings several other benefits - the quality of model architectures can be more precisely measured in the target environment, and user privacy can be better protected because there is no need to collect edge information. 
% Third, it is easier to maintain models when the edge environment changes.

% The most representative ones are neural architecture search (NAS) and model scaling (MS).
% NAS is usually performed on the cloud to search and train the appropriate model architectures \cite{}.
% However, most NAS approaches rely on target environment information collected from the edge devices, which may be impractical due to the diversity and dynamicality of edge environments and privacy concerns of collecting edge information.
The idea of adapting the model to the target device has been explored in both the mobile computing community \cite{legodnn,nestdnn} and the machine learning community \cite{network_slimming,skipnet}.
The mobile community is mainly focused on model scaling, \ie adjusting the model complexity to fulfill certain latency requirements, while ML research mainly aims to deal with different data distributions or hard/easy samples.
Model scaling approaches share a similar goal as ours, but prior work only shrinks the model size through pruning or quantization instead of changing the model architecture, which limits the opportunity to achieve optimal accuracy-latency tradeoffs.

% % Benefits of our method
% We propose to combine the advantages of NAS and MS by adapting neural architecture search to the target edge environments.
% Doing so brings several benefits. First, the quality of model architectures can be more precisely measured in the target environment.
% Second, user privacy can be protected because there is no need to collect edge information to cloud. 
% Third, it is easier to maintain models when the edge environment changes.

% % Challenges
% However, moving NAS to the edge faces several important challenges.
% \yuanchun{1. Unlike traditional NAS, training the model during searching is infeasible due to the limited training abilities. 2. Limited data makes it difficult to train from scratch. 3. Searching the large model architecture space is time-consuming on resource-limited edge devices.}
% % Shiqi: challenges might be coarse.

To this end, we introduce \name, an end-to-end system to generate models for diverse edge environments through post-deployment on-device neural architecture adaptation.
We focus on two related challenges in \name.
First, generating the search space of model architectures is non-trivial since the space must contain enough high-quality candidates that are suitable for different edge conditions.
Second, directly searching the optimal model architecture at the edge may be time-consuming due to the limited on-device resources.

\name addresses the above challenges by training a supernet once and letting the edge devices choose the satisfactory subnet on their own. The method can be divided into two stages, the on-cloud elastification and on-device search. % a two-stage approach with several edge-oriented optimizations. 

In the first stage, we design an on-cloud model elastification method to generate a high-quality search space for edge devices. Specifically, the elastification takes an arbitrary pretrained model as the input and converts it to a multi-path supernet by adding branches into the pretrained model, ensuring each path in the supernet is a valid and useful model. 
We introduce block-wise knowledge distillation to train the newly added branches, which consequently improves the quality of the subnets.
Our supernet offers millions of model variations with different structures, and the edge side only needs to find the best structure iteratively without additional training.

In the second stage, to improve the efficiency of architecture search and update at the edge, we systematically optimize the search process according to edge characteristics.
% The bottlenecks of searching the optimal at the edge are model evaluation and data loading. 
We first build a performance model on the device by profiling each block in the supernet, which guides the candidate selection during the search, therefore reducing the number of iterations needed to find the optimal model architecture.
Then we introduce a reuse-based model evaluation method, which caches intermediate features across model candidates to reduce the time required to evaluate the models in each iteration.
% The optimization includes a performance model-guided search strategy that reduces the number of iterations needed to find the optimal model architecture and a reuse-based model evaluation method that reduces the time required by each iteration.
% To address both problems, we improve the searching strategy by introducing feature sharing technique and initialization policy. These techniques accelerate evaluating subnets and reduce evaluation iterations, thus reduce the searching overhead at the edge greatly.

To evaluate our approach, we conduct experiments with various popular tasks (image classification, object detection, and semantic segmentation) and models (ResNet \cite{resnet}, MobileNetV2 \cite{mbv2}, EfficientNetV2 \cite{efficientnetv2}, etc.) on three edge devices including Jetson Nano, Android smartphone (Xiaomi 12), and edge server (NVIDIA 2080 Ti GPU).
We compare \name with several strong baselines including LegoDNN \cite{legodnn}, FlexDNN \cite{flexdnn}, SkipNet \cite{skipnet}, and Slimmable Neural Networks \cite{slim}.
The results have shown that our method can achieve significantly better accuracy-latency tradeoffs than state-of-the-art baselines. For example, our method can generate models that have 46.74\% higher accuracy than those produced by other methods with a latency budget of 60\%.
Meanwhile, the overhead of our method is minimal which only takes 13 GPU hours for elastification in the cloud and 2 minutes for adaptation on the edge server.

Our work makes the following technical contributions:
%  \sq{How about merge 2 and 3 together, since they are the essential techniques we proposed to facilitate post-deployment NAS.}
\begin{enumerate}
    \item We propose and develop the concept of on-device post-deployment neural architecture adaptation, and implement it with an end-to-end system.
    \item We introduce a pretraining-assisted model elastification method that can effectively and flexibly generate a model search space, as well as edge-tailored strategies to search the optimal model from the space and maintain it at runtime.
    % \item We introduce a performance model-guided search strategy and a reuse-based model evaluation strategy to accelerate the on-device model architecture search process.
    \item Our method achieves significantly better accuracy-latency tradeoffs than SOTA baselines according to experiments on various edge devices and common tasks. The tool and models will be open-sourced for edge AI developers to use.
\end{enumerate}

%% file: tex/background.tex
\section{Background and Motivation}
\label{section:background}

% \yuanchun{Ready for comments \& suggestions.}
% \renewcommand{\shortauthors}{Wen et al.}

\subsection{Current Practice and Related Work for Edge Model Generation}

% % the popularity of on-device DNN
% Deep learning has achieved remarkable progress in many areas including computer vision, natural language processing, programming language understanding, and computational biology. Many of its applications, such as object detection, machine translation, speech recognition, have the value to significantly enhance existing on-device services or even enable new applications.

Deploying deep neural networks (DNNs) at the edge is increasingly popular due to latency requirements and privacy concerns.
Since DNN models are mostly computationally heavy, deploying them to the edge usually has to consider two characteristics of edge devices. First, edge devices are mostly resource-constrained. As a result, there are already a lot of efforts on improving the performance of DNN models on edge devices, including optimizing the DNN inference framework on heterogeneous edge devices \cite{ulayer,MOSAIC,asymo}, designing lightweight model backbones \cite{mbv1, mbv2, mbv3, shufflenet, efficientnetv1, efficientnetv2} and compressing the models to be deployed \cite{deep_compression, haq, lqnet,bit_goes_down, int8}.

% These methods can accelerate model execution by better utilizing edge resources. The proposed method directly searches the optimal model on the edge. In the searching process, our method can collect information about the acceleration techniques on the edge, which will help us evaluate the subnets precisely. Consequently, our method is compatible with these edge-friendly model inference techniques.}

Besides the resource limitation, another major challenge of edge environments is the diversity - model developers usually need to deploy a certain model to thousands even millions of devices that are different from each other.
The deployed models are usually expected to meet a certain budget of latency while achieving higher accuracy, or achieve certain expected accuracy while minimizing the latency.
Thus, customizing the model for different target devices becomes a necessity.

% \subsection{Current Practice to Handle Edge Diversity}

% \yuanchun{Introduce the cloud-centric approaches. They can be called ``Pre-deployment model customization''}

The current practices to handle edge environment diversity are mostly cloud-based pre-deployment approaches, \ie the central server generates models for different edge devices before distributing them for deployment.
Since manually designing models for diverse edge environments is cumbersome, the common practice is to use automated model generation techniques.
NAS~\cite{mnasnet,single_path_one_shot,fbnet,netadapt,dna} is the most representative and widely-used model generation method, which searches for the optimal network architecture in a well-designed search space.
Most NAS methods require training the architectures during searching \cite{nas-rl, evolutionarysearch, mnasnet, darts-gradient-based}, which is very time-consuming (10,000+ GPU hours) when generating models for a large number of devices.
% The cloud generate models for different edge devices based on network performance modelling. The most representative method is Neural Architecture Search (NAS) \cite{mnasnet,single_path_one_shot,fbnet,netadapt, dna}. 
% Although this technique can achieve optimal accuracy given a resource budget after sufficient training and searching, it often assumes simple and ideal environments, and ignores the diversity and dynamic of edge devices. Besides, it often requires edge information and data collection, which may hurt the privacy of users. 
One-shot NAS \cite{proxylessnas, ofa, foxnas, sgnas_oneshot} is proposed to greatly reduce the training cost by allowing the candidate networks to share a common over-parameterized supernet.
Among them, several approaches also mention the concept of directly searching the architecture for target data and devices \cite{proxylessnas,foxnas}.
However, they require to collect much information from the edge devices to build accuracy and latency predictors, which are used to guide the search process in the cloud.

There are also several approaches proposed to scale models at the edge to provide a wider range of resource-accuracy trade-offs. 
Most of them apply structured pruning (or similar techniques) to generate various descendent models \cite{nestdnn, slim, us_slim, legodnn, network_slimming, filter_pruning, structured_pruning}, which can adjust the size of each network module without changing the architecture.
However, they have limited abilities to generate optimal models for diverse edge environments due to the restricted model space. Dynamic Neural Networks \cite{dynamic_neural_network_survey} are a type of DNN that support flexible inference based on the difficulty of input. When the input is easy, Dynamic Neural Networks can reduce the computation by skipping a set of blocks \cite{skipnet, blockdrop} or exiting from the middle layers \cite{early_exit, hapi, adaptive_inference, flexdnn}. However, this kind of work only considers dynamically adjusting the depth of DNN models, and they are not completely suitable for situations where latency budgets are strict. 
% On the model designing side, Neural Architecture Search \cite{oneforall,single_path_one_shot} and Dynamic Neural Networks \cite{skipnet,blockdrop,early_exit,dual_dynamic_inference} can provide multi-capacity models with different architectures, but they are incapable of on-device resource-aware scaling.
% Our work advances the model scaling research by enabling flexible architecture adaptation at the edge.
% Compared to pruning- or slimming-based scaling methods, we can greatly enrich the model varieties and hence generate more suitable models for diverse edge environments.
% \wh{Hardware-aware NAS technology can select optimal model for a specific device after sufficient training. However, traditional cloud-centric Neural Architecture methods \cite{mnasnet,proxylessnas,single_path_one_shot,ofa,fbnet} often need to collect device information to search for the optimal model on the server or to build latency predictors. Fox-NAS\cite{foxnas} introduces quantization-friendly methods, but still can not directly search for the optimal subnet on device. The main reason is the limited computation resources on edge devices. Our method can directly evaluate the performance of subnets by reducing the searching cost. As a result, our method does not need to connect device information to the server and performs better when the data distribution is different from the training data. }
% \yuanchun{merge the above two paragraphs}
% \setlength {\parskip} {0.0cm}

\subsection{Limitations of Current Practice}

We conduct several motivational studies to understand the limitations of the conventional model generation method.

We argue that the cloud-based pre-deployment model generation approaches underestimate the diversity of edge environments. We identify three types of diversity:

\begin{enumerate}
\item \textbf{Inter-device diversity.}
Edge devices are equipped with various types and grades of processors for DL inference, such as CPU, GPU, and AI accelerators. Even for devices with the same type of hardware, their conditions can be different. We measure the inference latency of two popular DNN models on four different mobile devices. As shown in Table~\ref{tab:bg:inter-device}, the inference latency of a model varies a lot on different devices.
% Thus, customizing the model for different devices is necessary to fully utilize the computation ability within certain latency budget.

\begin{table}
	\caption{Average latency (ms) of two DNN models on four mobile phones.}
	\vspace{-0.3cm}
	\centering
	\resizebox{.47\textwidth}{!}
    {
	\begin{tabular}{crr}
		\toprule
		%\multicolumn{2}{c}{Part}                   \\
		\cmidrule(r){1-2}
		Device    &  MobileNetV2   & ResNet50 \\
		\midrule
		XIAOMI 12 (Snapdragon 8 Gen 1) & 14.43 & 90.67\\
		HUAWEI nova 4 (HiSilicon KIRIN 970) & 53.05 & 372.22\\
		Google Pixel 2  (Snapdragon 835 64)   & 46.45       & 283.74\\
		Google Pixel 6 Pro (Google Tensor) & 31.29 & 144.21 \\
		\bottomrule
	\end{tabular}
	}
	\label{tab:bg:inter-device}
\end{table}   

\begin{table}
	\caption{Average latency (ms) of two DNN models under different conditions on NVIDIA 2080 Ti. The background processes use the same setting of the process in the ``normal'' condition, \ie continuous DNN inference with batch size 32 and CUDA version 11.3. In the ``CUDA version changed'' condition, we change the CUDA version to 10.1. In the ``different batch size'' condition, the batch size is set to 64.}
	\vspace{-0.3cm}
	\centering
	\begin{tabular}{crr}
		\toprule
		%\multicolumn{2}{c}{Part}                   \\
		% \cmidrule(r){1-2}
		 Condition   & MobileNetV2      & ResNet50 \\
		\midrule
		Normal                  & 13.35   &  33.79\\
		1 background process    & 14.37   &  50.36\\
		3 background processes  & 24.07   &  115.09\\
		CUDA version changed    & 13.69   &  35.89\\
		Different batch size    & 12.23   &  31.71 \\  
		\bottomrule
	\end{tabular}
	\label{tab:bg:intra-device}
\end{table}     

% \begin{table}
% 	\caption{Sample table title}
% 	\centering
% 	\begin{tabular}{lll}
% 		\toprule
% 		%\multicolumn{2}{c}{Part}                   \\
% 		\cmidrule(r){1-2}
% 		 Version   & MobileNetV2      & ResNet50 \\
% 		\midrule
% 		cuda 11.3 & 24.7   &  53.8\\
% 		cuda 10.1 & 20.7   &  56.3\\
% 		\bottomrule
% 	\end{tabular}
% 	\label{tab:process}
% \end{table}   

\item \textbf{Intra-device diversity.}
% \wh{
Even on the same device, the inference latency of a model may also be affected by various factors, including background processes, software versions, hardware aging, ambient temperature, etc. Table~\ref{tab:bg:intra-device} shows the non-neglectable impact of varying conditions on inference latency.
% So it is important for us to detect the changes of environment in real time and update the model.

\item \textbf{Data distribution diversity.}
NAS approaches need to search for the optimal architecture over a given dataset. However, edge devices are usually used in different locations and by different users, dealing with different data distributions \cite{personalizedfedlearning}. For example, some smart cameras are deployed in outdoor scenarios while some are indoor, and the common classes of objects in the scene may be different across devices.
\end{enumerate}

Such complex and ubiquitous diversity poses several difficulties for cloud-based model generation.
\textbf{First, tailoring models for diverse edge environments is a heavy task.}
To generate optimal model architecture for each edge environment, the current practice requires repeating the search process for all types of environments and maintaining them in the cloud.
% However, we argue that the edge environments are too diverse for the developers to deal with in fine granularity. 
The required manual and computational effort are determined by the granularity of edge environment diversity to consider, which might be burdensome if the developers want to achieve optimal latency-accuracy tradeoffs on more devices.
Meanwhile, handling the dynamicity of the edge environment is even more difficult since it requires frequent communications with each edge device and rapid reactive model updating in the cloud.
% We argue that the required effort might be a heavy burden due to the environment diversity.

% Taking all these conditions into consideration in model generation is difficult. Thus, existing cloud-based model customization approaches have to apply corased-grained model customization, which has to sacrifice accuracy on some devices to meet the latency budget.

\begin{figure}
    \centering
    \includegraphics[width=8.5cm]{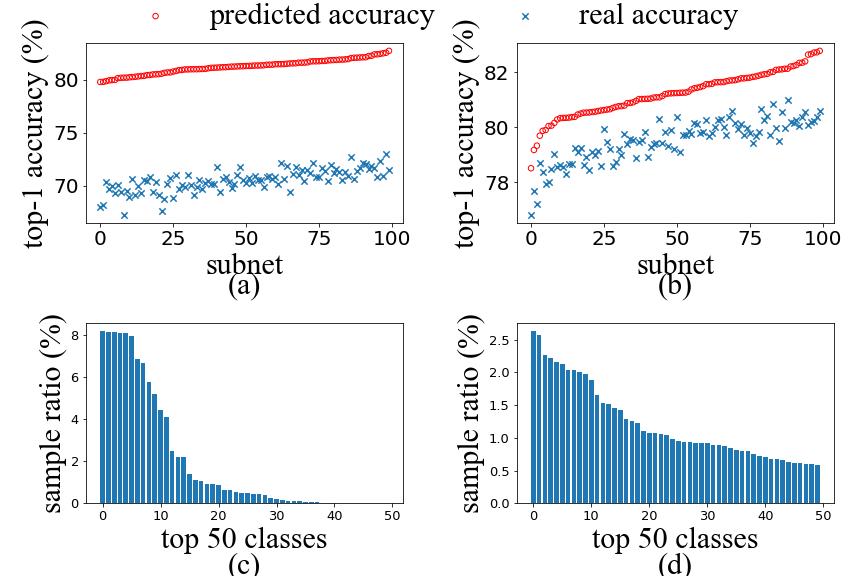}
	\vspace{-0.5cm}
    \caption{Performance of cloud-trained accuracy predictor on distribution-shifted edge data. The edge data is simulated with Dirichlet distributions with (a) $\alpha=0.005$ and (b) $\alpha=0.1$. The sample ratios of top-50 classes are shown in (c) and (d).}
	%  \yuanchun{normalize the sample ratios; increase font size.} \sq{fix the position of legends}}
    \setlength {\parskip} {-0.2cm}
    \label{fig:predictors}
\end{figure}

\textbf{Second, modeling the edge environment may also be difficult.} A necessary step in the cloud-based model generation is to estimate the performance of the candidate model, such that the model architecture can be optimized according to the target hardware and data. For example, existing NAS methods are usually based on accuracy and latency predictors \cite{oneforall}. Building the predictors requires collecting intensive edge information, which is not easy, especially for the accuracy predictor that depends on the potentially private edge data.
The compromise solution is to use a unified accuracy predictor for different edge devices \cite{ofa}.
% A common constraint of above two types of works is that they both depend on the predictors built on cloud to estimate the accuracy of subnets on the edge. 
However, the unified accuracy predictor may not perform well for edge devices with data distribution shifts. As shown in Figure \ref{fig:predictors}, the accuracy values and rankings of candidate models predicted by the once-for-all accuracy predictor \cite{ofa} are both different from the ground truth, which indicates that the predictors can be unreliable on edge distributions, leading to sub-optimal model generation.
% However, data distributions at edge are often different from the cloud\cite{personalizedfedlearning}, thus the predictors can be biased on devices, leading to sub-optimal search results. Figure \ref{fig:predictors} shows the performance of accuracy once-for-all accuracy predictor \cite{ofa} on Non-IID data distributions sampled in validation set of ImageNet2012\cite{imagenet}. The real accuracy of subnets waves with the predicted accuracy goes up, which indicates that the predictors can be unreliable on edge distributions, leading to sub-optimal search results. 

\subsection{Post-deployment Neural Architecture Generation: Goal and Challenges}

The limitations of existing edge model generation methods motivate us to think, can we directly search for the optimal neural architecture on the edge device after deployment?

Doing so brings several key advantages. Unlike traditional on-cloud NAS which has to estimate the performance of subnets, edge devices can directly evaluate the performance of a given model architecture natively, which is more precise. Besides, searching on the device is a plug-and-play process and does not need to collect edge information to the cloud, bringing the benefits of protecting user privacy and reducing the computation overhead of the cloud.
% \sq{Data is from edge, but where the ground truth comes from?} \yuanchun{probably add a paragraph in the discussion section for this}

On the other hand, finding the optimal model architecture directly at the edge is challenging.
\textbf{First, generating the model search space for edge devices is difficult}. The search space should be flexibly and easily customizable to support diverse edge applications and different ranges of target devices. Meanwhile, since the training abilities of edge devices are usually weak, the search space should contain high-quality candidate models that can be used in different edge environments with minimal (or even no) further tuning.
% the edge applications are varied, thus the NAS process should be universal.
\textbf{Second, the model search process can be time-consuming at the edge.} Existing architecture search methods require either training a lot of candidate models or repeatedly evaluating the performance of the candidates. Both are very heavy for the edge devices because of the limited computing resources and tight deadline of model initialization.
% Training the candidate models would be even harder (if not impossible).
Dynamically updating the model according to environment changes is even more time-sensitive.
% Finally, update overhead of the subnets should be minimal because the environment of edge device is dynamic. 
% \yuanchun{Limited computing resource; Limited training data.}

%% file: tex/approach.tex
\section{\name Overview}
\label{sec:approach}

% \yuanchun{ready for comments/suggestions}
% \sq{How about changing the title to \textit{AdaptiveNet}}.

To solve the aforementioned challenges and realize the vision of post-deployment model generation, we introduce \name, an on-device neural architecture adaptation approach for diverse edge environments. To the best of our knowledge, \name is the first end-to-end system to enable on-device architecture adaptation.
% \sq{can we claim the first of its kind here?}
% \yuanchun{Describe the core idea with one or two sentences.}

% \subsection{Overview}

\begin{figure*}
    \centering
    \includegraphics[width=16.4cm]{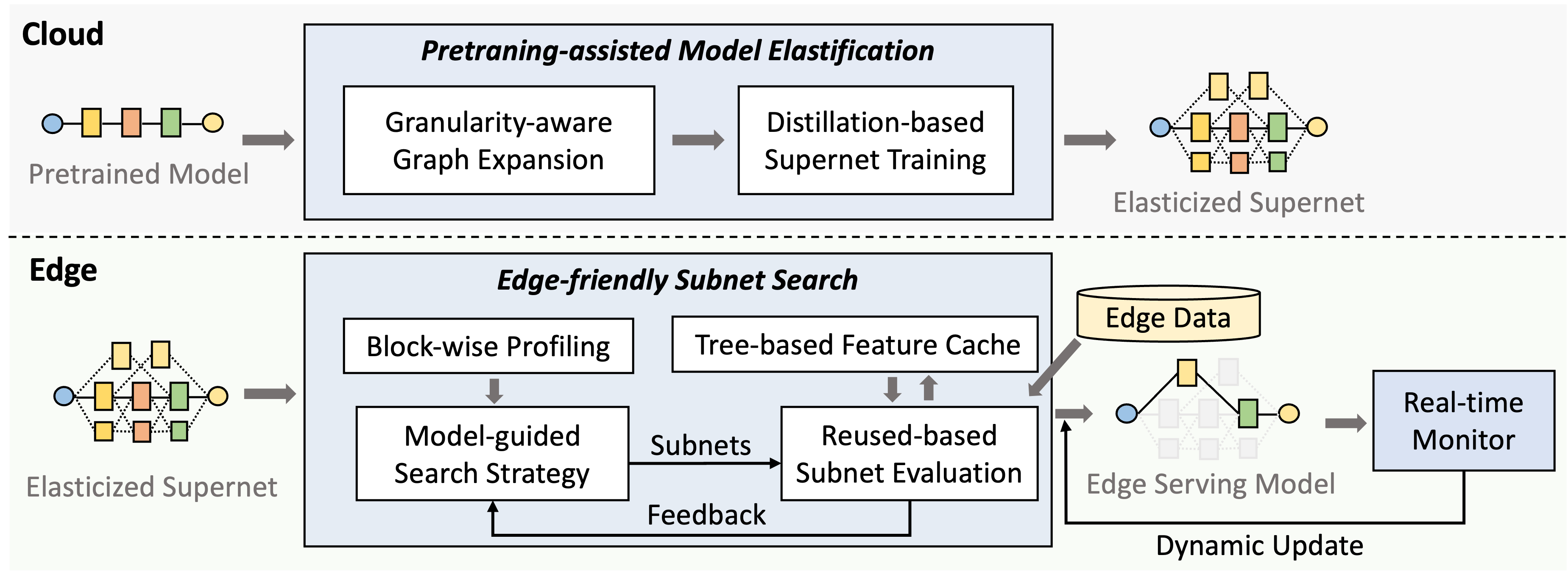}
    \vspace{-0.3cm}
    \caption{The architecture overview of \name.}
    \vspace{-0.3cm}
    \label{fig:system_architecture}
\end{figure*}

The main idea of \name is to generate high-quality model search spaces based on developer-specified pretrained networks through modular expansion and distillation, and efficiently search for the optimal architecture on the target device guided by performance modeling.
Figure \ref{fig:system_architecture} illustrates the architecture of \name, which includes an on-cloud model elastification and an on-device subnet search. 

Our model elastification is efficient by leveraging the guidance of a developer-specified pretrained model. It mainly consists of a granularity-aware graph expansion step and a distillation-based training step.
Given an arbitrary pretrained network, we first discover the repeating basic blocks and determine the replaceable paths in the computational graph. Then we add optional branches to the model to extend it into a supernet. The added branches include layers that can replace multiple original layers, or structured-pruned layers that reduce the computational cost of individual layers. Each path from the input to the output in the graph is a valid subnet, which consists of both original and newly added modules.
%  and the number of optional branches in a subnet decides the latency reduction. 

The supernet obtained by graph expansion contains a large variety of architectures with different levels of computational complexity. We then further improve the quality of each candidate architecture in the supernet through training, so that on-device training can be avoided to save computation cost.
Since our supernet is generated from a pretrained model, we use branch-wise distillation to efficiently train the newly-added branches to mimic the original branches.
The distillation is followed by a whole-graph fine-tuning to further improve the overall accuracy of the subnets.
With all these techniques, the supernet would contain high-quality subnets that can fit in different edge environments, and it is deployed to the edge devices for further adaptation.
% the branch-wise distillation phase (Section 3.2.2), \name trains optional branches guided by the original pre-trained model. We introduce feature map knowledge distillation to speed up the convergence of training. We uniformly sample a single subnet consisting of several optional and original branches for data batch, and freeze the wights of original branches when distilling. 

% After the distillation phase, we further train the optional branches using image-label pairs \sq{using labeled data?} The training method is still uniformly sampling subnets, and we use Cross Entropy as the loss function. This phase can further improve the accuracy of subnets (Figure \ref{fig:distillation_gain}). \sq{why this step is necessary?}

The on-device subnet search stage aims to find the most appropriate subnet (that can achieve the highest accuracy within the latency budget) on resource-limited edge devices.
We first build a latency model by profiling the blocks in the supernet to precisely estimate the latency of subnets in the native environment.
Based on the latency model, we design a search strategy that initializes a set of promising candidate models and iteratively mutates the candidates around the latency budget.
The search efficiency is further improved by reusing the common intermediate features during candidate model evaluation.
The optimal model is also adaptively updated by the runtime monitor to handle environment dynamicity.
% the on-device subnet search stage consists of three modules, module profiling, subnet searching and runtime update. In the module profiling phase, \name performs layer-wise inference to measure the latency of each branch. After profiling the latency models, \name searches the optimal subnet using the local data \sq{where is the labeled data from?}. To speed up the search process, \name introduces feature-sharing strategy (Section 3.3.2) and estimate initialization strategy (Section 3.3.3). After deploying the optimal model, \name continuously monitors the latency of the model, and adjusts the DNN model when the latency changes. 

% \section{On-cloud Model Elastification}
\section{Elastification on Cloud}
\label{sec:elastification}

% \yuanchun{Let's use Elastification instead of Elasticizing}

% \sq{Made some modifications, please check and merge.}

The input of the model elastification stage is a developer-specified pretrained model, similar to the common scenarios in edge AI deployment. The pretrained model is determined as the best-performing model that can fulfill (or slightly exceed) the latency budget on the highest-end target device.

The goal of elastification is to convert the given pretrained model into a \emph{supernet}, by expanding alternative basic blocks, making connections between them, and letting each path in the supernet (namely \emph{subnet}) behave correctly.
% Thus, a supernet contains a lot of unique paths between the input and output nodes, namely \emph{subnets}. We train the supernet to let every subnet behave correctly with an acceptable accuracy.
In this way, the supernet is granted with the \emph{elasticity}: each subnet has the discriminative performance characteristics in terms of inference latency and accuracy. The supernet is then deployed onto the edge, where the particular edge device can search for the most suitable subnet according to its own hardware capacities and data distribution.

There are two main problems to solve in elastification. The first is how to generate the supernet architecture. Although prior work \cite{ofa} has discussed hand-crafted supernets for certain models, it is still an open question to automatically generate supernets based on an arbitrary pretrained model, especially considering the diversity of DNNs.
% The elastification process should be automatically applied to different kinds of computing graphs. 
Another problem is how to train the subnets in the supernet to improve their quality. A supernet typically contains millions of subnets, thus training them separately is time-consuming.
% A block in the supernet might be shared by multiple subnets, thus training on one subnet might influence the training of another.
We propose two techniques to address these problems accordingly.

% \sqedit{To this end, we propose two techniques accordingly, the automatic supernet generation and the distillation-based supernet training. Next we would describe the details.}

% Elasticizing means adding alternative branches into a pretrained model and make it a supernet which contains a large number of subnets. We train the supernet so that the subnets in it can achieve acceptable accuracy. Note that this is not a easy task, because the computing graph of DNNs can be very diverse, and our elastification process should be automatically applied to different kinds of graphs. Moreover, the subnets can influence each other since they share some of weights,so training the supernet can be difficult.

\subsection{Granularity-aware Graph Expansion}
\label{sec:supernet-generation}

% \yuanchun{Describe how we add branches to generate the supernet.}

\begin{figure}
    \centering
    \includegraphics[width=8cm]{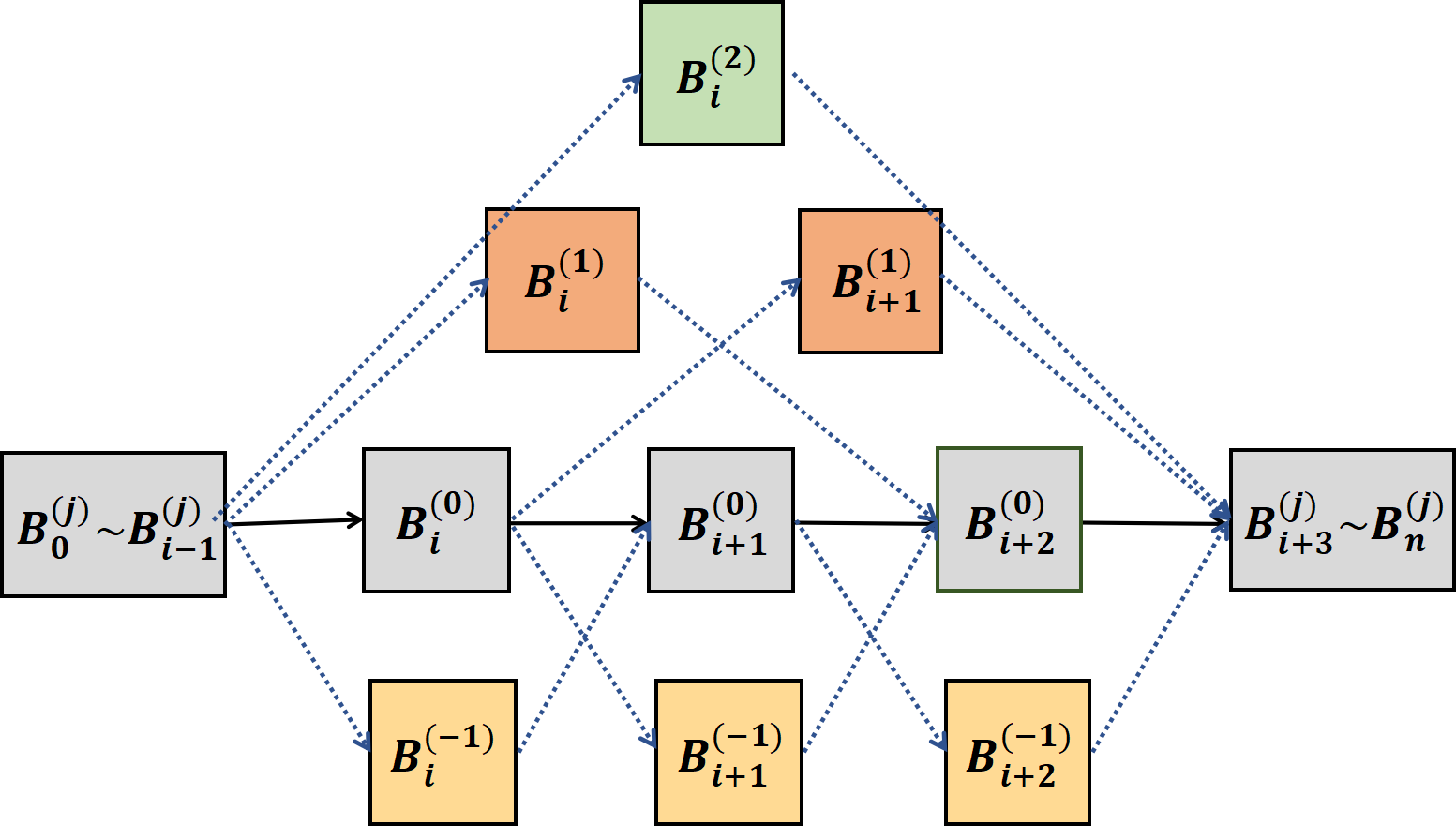}
    % \vspace{-0.2cm}
    \caption{Supernet architecture in \name.}
    % \vspace{-0.3cm}
    \label{fig:supernet}
\end{figure}

Let $\mathcal{N}$ denote the pretrained model we want to elasticize. The first step is to analyze the computational graph of $\mathcal{N}$ to determine how it can be expanded.
We call the smallest unit in the graph that be replaced as a \emph{basic block}, and the block partitioning is determined by the following principles.
First, the size of blocks determines the subnet search space size and the granularity of how the latency can be controlled. Thus, we limit the block parameter size to no more than $\gamma \cdot P_0$ where $\gamma$ is a parameter to control the granularity and $P_0$ is the parameter size of the original model $N$.
Second, the blocks should not span fusion layers. For example, Conv and ReLU can be fused in most inference frameworks \cite{layerfusion}. 
% We build a whitelist to .
Third, each basic block should be single-input and single-output in the original model graph.
Following these principles, we can represent the supernet as a set of connected basic blocks $\mathcal{N} = graph \{ \mathcal{B}^{(0)}, C\}$, where $\mathcal{B}^{(0)}$ is the set of blocks and $C$ denotes the connections between them.

Next, we generate the supernet graph $\mathcal{S}$ by expanding the graph of the pretrained model $\mathcal{N}$ through adding alternative blocks and connections. Particularly, we consider two expanding strategies including merging and shrinking, as shown in Figure \ref{fig:supernet}.
First, we add merged blocks $B_i^{(j)}$ that can replace multiple basic blocks ($j>0$ represents the number of reduced blocks in the replacement).
% $B_{i...j}^{(0)} = \{ B_i^{(0)}, B_{i+1}^{(0)}, ..., B_{i+j}^{(0)}\}$ \yuanchun{what does this formula mean?}.
Suppose $\{ B_i^{(0)}, B_{i+1}^{(0)}, ..., B_{i+j}^{(0)} \}$ are the basic blocks in $\mathcal{N}$ that can be replaced by $B_i^{(j)}$, The input shape of $B_i^{(j)}$ is the same as $B_i^{(0)}$, and output shape is the same as $B_{i+j}^{(0)}$. The parameter size of $B_i^{(j)}$ is determined by the largest among the replaced blocks.
Second, like traditional model scaling approaches \cite{legodnn,nestdnn}, we also add different levels of shrunk blocks $\mathcal{B}_{i}^{(-1)}, \mathcal{B}_{i}^{(-2)}, ...$ for each basic block $\mathcal{B}_{i}^{(0)} \in \mathcal{B}^{(0)}$ by reducing its size with structured pruning and network slimming techniques \cite{slim,us_slim,filter_pruning}.
The granularity of merged blocks and pruned blocks can be balanced to control the size of subnet search space.
% \sq{consider changing the block color in fig.4}

% After analysis, we expand the computational graph of the pretrained model to make it a supernet. We define the expansion process as elasticizing, which is shown in \ref{fig:supernet}. The optional branches include two types of blocks, the first one is denoted as $B_i^{(j)}(j>0)$, which can replace $B_{i...j}^{(0)}=\{B_i^{(0)}, B_{i+1}^{(0)}, ...B_{i+j}^{(0)}\}$. Its input dimension is the same with $B_i^{(0)}$, and output dimension is the same with $B_{i+j}^{(0)}$. The width and kernel size of an optional branch is determined by the largest among the blocks it replaces, because we want$B_i^{(j)}$ to extract as much information as possible. Note that not all the blocks can be replaced by optional blocks. During the graph expansion process, we detect those blocks with multi connections, and forbid them from being elasticized, since this may change the DNN architecture.

% The other type of optional branches are the descendant blocks of $B_i^{(0)}$ denoted as $B_i^{(j)}(j<0)$. The descendant blocks can be generated by structured pruning, which proves that \name is compatible with former network slimming methods\cite{slim}\cite{us_slim}\cite{legodnn}.

Compared to the existing model scaling methods \cite{legodnn,nestdnn,slim}, our supernet has higher elasticity because it allows the subnets to have different architectures, rather than just different sizes of the same architecture. This is also the reason why NAS outperforms other model generation techniques on the server side \cite{efficientnetv1, ofa}.
% due to two reasons. First, rather than one simple block, we conduct the expanding on multiple basic units. Second, we do not only expand the descendant blocks but the fused blocks as well, which would largely reduce the inference latency.

% that only add alternative pruned blocks for one simple block, we add alternative blocks for two or more blocks. Experiences show that compressing two or three blocks into one blocks can reduce inference latency greatly compared to simply pruning one block. When a batch of data passes through the supernet, we can control the path for it by scheduling the alternative blocks it passes. By doing so, we can adjust the latency of a model by changing the subnet path.

\subsection{Distillation-based Supernet Training}
\label{sec:distillation}

% \subsubsection{Branch-wise Distillation}
Next, we need to train the generated supernet to improve the quality of its subnets, so that the subnet can be directly used at the edge without further training. We achieve efficient and effective training by fully utilizing the supernet. The whole training process includes a branch distillation phase and a whole-model tuning phase.
% The training goal is to make the achieved accuracy of every subnet be not lower than the accuracy achieved by the origin pre-trained model $\mathcal{N}$. We propose the two-stage supernet training.

\begin{figure}
    \centering
    \includegraphics[width=8cm]{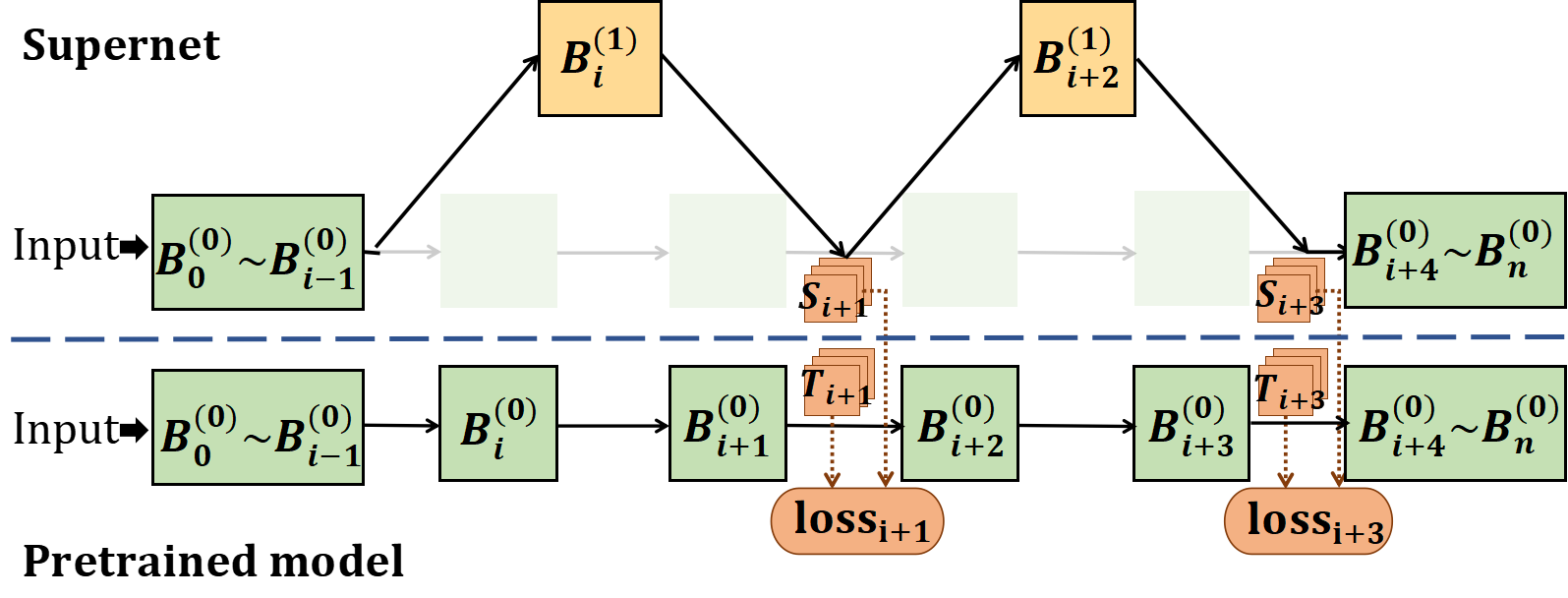}
    \vspace{-0.2cm}
    \caption{Illustration of the branch-wise distillation phase.}
    \vspace{-0.3cm}
    %  \sq{the figure is blur, consider using the PDF version, as well as all other figures.}
    \label{fig:distillation}
\end{figure}

\textbf{Branch-wise distillation}. In this phase, we first freeze the weights of $B_i^0$ so that the accuracy of the original pretrained model is preserved. Then we adopt feature-based knowledge distillation~\cite{feature_distill} to let the added blocks imitate their corresponding original blocks. As illustrated in Figure \ref{fig:distillation}, in each iteration, we randomly sample a subnet from the supernet and use $\mathcal{N}$ as the teacher model to train the new branches in the subnet. Specifically, let $S_{i}$ denote the output feature map of a newly added block $B_i^{(j)}$ and $T_i$ denotes the output feature map of the last block that $B_i^{(j)}$ replaces, we use the L2 distance as the distillation loss. The loss function is
\begin{equation}
    \mathcal{L}_{distillation}=\frac{1}{M} \sum_{i=1}^{M} \left \| T_i-S_i \right \| _2 ^2,
    \label{distillation}
\end{equation}
where M denote the number of new blocks in the sampled subnet. 
With enough iterations applied, all new blocks in the supernet will be trained multiple times to improve their individual quality.
Since we only train the new blocks and use the feature maps of the pretrained model as strong supervision, the distillation process is efficient and easy to converge.
% \sq{how random sampling work, and how it would influence the converge.}

% \wh{Randomly sampling is an efficient way to train supermodel and is widely applied to NAS and MS techniques \cite{single_path_one_shot, dna, mutualnet, ofa}. LegoDNN \cite{legodnn} introduces block-wise retraining to generate input and output of each descendent block and trains it independently. We use randomly sampling technique instead of block-retraining because optional blocks can learn to function in different subnets rather than only learn from the original DNN model. }

% illustrates the details of supernet distilling. In order to preserve the accuracy of original pretrained model, we only update the weights of $B_i^{(j)}$ during training. As a result, the highest accuracy of the supernet won't be lower than the accuracy of $\mathcal{N}$. Therefore, our training goal is to let the optional branches imitate the original blocks they replaces.

% According to our training goal, we introduce a two stage training technology to train the supernet. In the first stage, we use distillation method to teach $B_i^{(j)}$ to learn from $B_{i...j}^{(0)}$. For each batch of data, we randomly sample a subnet from the supernet, and use $\mathcal{N}$ to teach it, as shown in \ref{fig:distillation}. Let  $S_{i}$ denote the output feature maps of i-th newly added optional block and $T_i$ denote the output feature maps of the last block the optional block replaces, we take their L2 distance as the distillation loss. Therefore, the loss function can be written as:
% \begin{equation}
%     \mathcal{L}_{distillation}=\frac{1}{M} \sum_{i=1}^{M} \left \| T_i-S_i \right \| _2 ^2 
%     \label{distillation}
% \end{equation}
% where M denote the number of optional blocks in the sampled subnet. Since we only train the optional blocks and use the feature maps of the pretrained model to teach the subnets, the distillation process's overhead is little and only costs several epochs to converge. 

% \subsubsection{Further Training.}
\textbf{Further tuning.}
We further train the supernet using labelled data to improve the end-to-end quality of the subnets. In each step of tuning, we randomly sample a subnet $B_i^{(j)}$, forward a batch of samples, compute the Cross-Entropy~\cite{cross_entropy} loss between the subnet outputs and the labels, and update the parameters of the added blocks in $B_i^{(j)}$ via gradient descent.
The performance of the supernet is measured by sampling a new set of random subnets and testing each of them on validation data. We use the \emph{latency-range accuracy} as the training progress indicator, which records the average accuracy achieved by subnets in each latency range.
This phase starts from the distilled supernet, and thus the learning rates are relatively small and the convergence is fast.

\textbf{Notes on design rationale.} Each phase in our design is indispensable to ensure training efficiency and effectiveness. Using distillation only will lead to suboptimal final accuracy, and direct training will significantly slow down the convergence. Merging the two phases together is also not desirable since it will make the loss design and training more difficult.
The experimental comparison can be found in Section~\ref{eval:training}.
We also note that our method does not modify the parameters of the pretrained model, so it guarantees that the latency-accuracy tradeoffs will be better than or at least equal to the pretrained model.
% Compared to merging the distilling and further tuning stages together by computing the loss as $\mathcal{L}_{total}=\mathcal{L}_{distillation}+\alpha \cdot \mathcal{L}_{cross-entropy}$, our two-stage training method is hyperparameter-free and converges faster. \sqedit{According to our evaluation, the further tuning stage can bring up to 10\% accuracy improvement.} \sq{Again, how random sampling impacts the convergence. And why it is necessary?}

% We find out that further training can improve the accuracy of subnets significantly (nearly 10$\%$ percent top-1 accuracy on ImageNet2012\cite{imagenet} as showm in Figure \ref{fig:distillation_gain}).

% \section{On-device Architecture Adaptation}
\section{Adaptation on Edge}
\label{sec:adaptation}

The supernet generated by model elastification is uniformly deployed to different edge devices, but it is not directly usable since each edge device has different characteristics and requirements.
Thus, we further introduce the on-device adaptation stage to obtain the optimal architecture adaptively in the target environment by searching the subnet space.
Such a search process is similar to traditional on-cloud NAS but has a higher requirement for efficiency.

% \textbf{Bottleneck Analysis.}
% \yuanchun{We first analyze the performance bottleneck of searching the subnets on the device.}
% Compared with Cloud-centric Neural Architecture Search approaches, On-device NAS faces the difficulty of limited computation resources.
% For example, searching for the best subnet from a ResNet50-based supernet on Jetson Nano with 4 GB memory can cost 10 GPU hours. 
According to our analysis, using a normal search method as in NAS can cost more than 10 hours on edge devices.
Most of the searching time is spent on evaluating the subnets.
% According to our observation, the evaluation process takes up \xx of the whole searching overhead.
This is because we have to perform model inference hundreds of times to get the accuracy of candidate models in each search iteration and use the accuracy results to guide the next search iteration.
% \sq{the validation overhead is highly related to the size of validation set, it is not fixed, so a naive question here is how about reducing the validation set?}
% one single subnet using local data. During one iteration of evolutionary search, we need to evaluate hundreds of subnets, thus evaluating all the subnets may take several GPU hours. 
To reduce the searching overhead, we introduce a latency model-guided search strategy and a reuse-based model evaluation method.

% \yuanchun{Based on the above analysis, we introduce three techniques to improve the efficiency of the adaptation process on edge devices.}

% \yuanchun{This para does not belong here. Should be included in next subsection.} To cut down the searching overhead, Cloud-centric NAS techniques build accuracy-predictors to estimate the performance of models on the edge. However, the data distribution on the edge is usually different from the training data, which may make the predictor biased. Therefore, \name directly evaluates subnets on the edge to get their precise accuracy. To solve the difficulty of search overhead, we introduce feature-sharing search strategy and initialization strategy to reduce the evaluation overhead and evaluation times respectively. The whole searching strategy is summarized in Algorithm\ref{nas}. 

\subsection{Model-guided Search Strategy}
\label{sec:on-edge:search}

% \yuanchun{this subsubsection should be largely extended. At least it should include one paragraph on how to build the latency model, one para for the fast model initialization, and one para for the nearby mutation strategy.}

We first optimize the search strategy to find the optimal model architecture (\ie the architecture that can achieve the highest accuracy within the latency budget) with fewer iterations. The core idea of the optimization is to prune the search with the guidance of a natively-built latency model.

Formally, suppose $T_{budget}$ denotes the latency budget in the target environment and $D_{edge}$ is the edge dataset.
Our goal is to find a subnet $\mathcal{N}' = \{ B_{i_1}^{(j_1)}, B_{i_2}^{(j_2)}, ..., B_{i_n}^{(j_n)} \}$ from the supernet $\mathcal{S}$ whose latency $T(\mathcal{N}') \leq T_{budget}$ and accuracy $Acc(\mathcal{N}', D_{edge})$ is optimal.
During the search, the accuracy of the candidate model is directly measured on the edge dataset $D_{edge}$, and the latency is computed with a latency model.

The latency model is a table $\mathcal{T}=\{T_i^{(j)}\}$ where $T_i^{(j)}$ is the latency of basic block $B_i^{(j)}$ in the supernet.
The block latency is precisely measured on the device through profiling after deployment. Note that this process is quick (within seconds) because the number of basic blocks is small (much smaller than the number of subnets).
Our supernet generation strategy (Section~\ref{sec:supernet-generation}) ensures that the basic blocks will not be fused, thus the latency of a chosen subnet is the sum of all its blocks, \ie $T(\mathcal{N}') =\sum_{k=0}^{n} T_{i_k}^{(j_k)}$.
Note that we use the latency model to compute the latency rather than directly measure it because end-to-end latency measurement under the actual model operating condition is time-consuming.

The subnet search process contains two main steps, including candidate initialization and candidate mutation, where the initialization step produces a set of seed subnets and each mutation step changes the subnets iteratively to better fit the target environment.
Both the initialization and mutation are customized with the latency model in our approach.
% , we build a branch scheduler to guide the generation of subnets and reduce the search space. \sq{confused, why subnets are generated here?} 
Our experiments show that the optimal subnets are often near the latency budget.
Therefore, the initialization and mutation are designed to keep the search of candidates near the budget.

% given the latency table and budget, branch scheduler generates subnets around the latency budget as the initial basic solution by iteratively generating and filtering subnets.
% For example, let $t_k$ and $T$ denote the latency of k-th subnet and the latency budget respectively, the initial basic solution should be met with the constraint of $t_k\in[T-\Delta T, T+\Delta T]$ ($\Delta T$ controls the variation range of latency, $T/10$ as an example). 
% \sq{not clear. How the search is performed? why delta T is needed? The basic search process is to enumerate all possible subnets and filter out the subnets from T-t to T+t ?}

Specifically, we design two supporting functions $NearbyInit$ and $NearbyMutate$. $NearbyInit$ generates the initial candidate subnets by randomly sampling a group of subnets whose latencies lie in the range of $[T_{budget} - \Delta T, T_{budget} + \Delta T]$. The models with latency higher than $T_{budget}$ are unlikely to be useful at the current moment, but they may be used later to handle dynamic environment change (see Section~\ref{sec:on-edge:update}).
Next, when we change the candidate subnets in each iteration, we first randomly mutate a subnet by replacing a branch in it. If the latency of the subnet after mutation is out of $[T_{budget} - \Delta T, T_{budget} + \Delta T]$, we iterate the rest branches in the supernet and find the best alternative branches that can reduce the latency change.

Both the initialization and mutation strategies are graph operations that do not involve heavy computation, but they can significantly reduce the evaluation overhead by improving search efficiency.
Meanwhile, the model-guided initialization and mutation can be integrated into most standard search algorithms including evolutionary search \cite{evolutionarysearch} and simulated annealing \cite{foxnas}.
Algorithm~\ref{nas} shows the subnet search strategy based on evolutionary search.
\begin{algorithm}
\footnotesize
	\caption{Edge-friendly optimal subnet search}
	\label{nas}
	\begin{algorithmic}[1]  %1表示每隔一行编号	
		\Require Elasticized supernet $S$, Local data $D$, Latency budget $T$.
		\Ensure  Optimal subnet $M$. 

        \Function{Main}{}:
        \State $candidates \leftarrow NearbyInit(S, T)$; // Section~\ref{sec:on-edge:search}
        \State $LatencyTable \leftarrow$ Block-wise latency table of $S$;
        \State $root \leftarrow BuildTree(candidates, LatencyTable)$;
        \State $DFS-Evaluate(D, root)$;
        \For{$i=0;i<search\_times;i+=1$};
        \State $candidates \leftarrow NearbyMutate(candidates)$; // Section~\ref{sec:on-edge:search}
        \State $DFS-Evaluate(D, BuildTree(candidates, LatencyTable))$;
        \State Record best candidate;
        \EndFor\\
        \Return Optimal subnet $M$;
        \EndFunction\\
		
		\Function{DFS-Evaluate}{$D,root$}:
		\If {$root.child == 0$}
		\State return Evaluate $root$;
		\Else
		    \For{each $child$ of $root$}
		    \State $PrefixFeature \leftarrow child.prefix.forward(D)$;
		    \State Save $PrefixFeature$ in cache;
		    \State $DFS-Evaluate(PrefixFeature, child)$;
		    \State Release $PrefixFeature$;
		    \EndFor
		\EndIf
        \EndFunction\\

		\Function{BuildTree}{$candidates,LatencyTable$}:

		\State Get $prefixes$ shared by $candidates$;
	%	\State Compute the importance of prefixes by latency-reduction * multiplexing-rate;
		\For {$prefix\in prefixes$}
		\State Get $prefix.latency$ from $LatencyTable$;
		\State $prefix.importance \leftarrow prefix.latency\times PrefixingRate$;
% 		\Statex $\quad \quad \quad \quad \quad prefix.latency\times prefix \_ multiplexing \_ rate$;
		\EndFor
		\State Delete less important prefixes if one subnet has several prefixes;
		\State Return the tree of subnets based on $subnet.prefix$;
		\EndFunction
	\end{algorithmic}
\end{algorithm}

\subsection{Reuse-based Model Evaluation}
\label{sec:on-edge:evaluate}

% \yuanchun{title changed from `Feature-sharing Search Strategy' to `Reuse-based Model Evalutation'. Pls change accordingly.}
% \sq{My biggest concern here is, how the evolutionary algorithm is used here? how it is couple with this reuse-based model evaluation?}

\begin{figure}
    \centering
    \includegraphics[width=8cm]{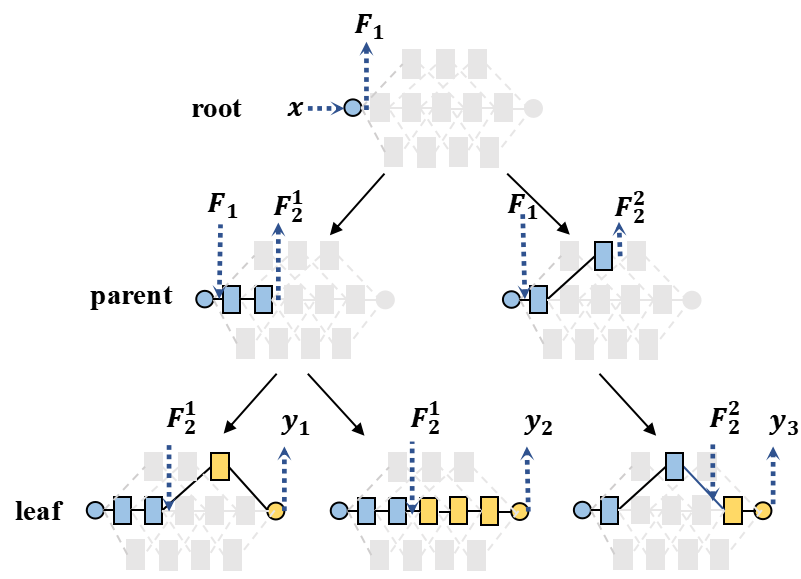}
    % \vspace{-0.2cm}
    \caption{Illustration of subnets tree. Prefix layers shown in blue, subsequent in yellow. $\bm{x, F_i^j, y_i}$ represent input, shared features, output respectively.}
    \label{fig:tree}
\end{figure}

While the model-guided search strategy reduces the required iterations, the model evaluation overhead in each iteration is still high.
% We try to further reduce this overhead by reusing the 
In each iteration, we usually need to evaluate hundreds of candidate subnets with the edge data to find the most accurate ones. The candidate subnets usually share common prefix substructures, so we have the opportunity to save time by reusing common intermediate features across subnets.
% Our Feature-sharing Search Strategy is based on Evolutionary algorithm, and can also be applied to other NAS algorithms. It is designed to save and reuse the important features that can be used in several subnets. During one iteration of evolutionary search, we have to evaluate hundreds of subnets, which share some common prefix computational graphs. 
For example, let $\mathcal{N}'_1 = G_{prefix} \cup G_1$ and $\mathcal{N}'_2 = G_{prefix} \cup G_2$ denote two different subnets and $\mathcal{N}'_1$ is evaluated before $\mathcal{N}'_2$,
% $S_1 = G_{prefix} \cup G_1$ , $S_2=G_{prefix} \cup G_2$, 
% given a batch of data $D$, 
during the evaluation of $\mathcal{N}'_1$, the output feature of $G_{prefix}(D_{edge})$ can be saved and reused when evaluating $S_2$, which saves the inference cost of computing $G_{prefix}(D)$.
% \sq{how about changing the symbol 'prefix', it seems too long for an inline math equation.}

However, saving all the common features during model evaluation is infeasible because it will take too much memory. Thus, we can only save part of the common features and improve the reuse ratio as much as possible. In order to achieve this, we introduce a tree-based feature cache to schedule the evaluation. The leaf nodes and non-leaf nodes in the tree represent subnets and common prefix substructures respectively. The leaf nodes (subnets) sharing the same parent node have the same prefix substructure represented by the parent node. And two non-leaf nodes with the same parent node have the same smaller prefix substructure. 
% The root of the tree represents the common prefix shared by all subnets in the tree.

After building the feature cache tree, we evaluate all the subnets in the depth-first order. When we traverse to node $N$, we cache the output feature of that node in memory. Then, when evaluating the descendant leaf nodes (subnets) of $N$, we can reuse the cached feature. After evaluating all descendant leaf nodes (subnets) of $N$, we can release the feature from memory since it won't be reused by later subnets. As a result, the number of saved features is no more than the depth of the tree, and we can adjust the depth of the tree to control the size of the feature cache.

Another problem of model evaluation is that testing the models one by one may lead to too frequent data I/O operations. Thus, we adopt \emph{batch-wise model group evaluation}, \ie loading a batch of data and evaluating all candidate subnets using the batch. The performance of the subnets is the average of them on all batches. 
% The depth or the tree of subnets is determined by the cache memory, because we save the features in the cache for fast access.
% \begin{figure}
%     \centering
%     \begin{subfigure}{0.47\textwidth}
%         \centering
%         \includegraphics[width=8.3cm]{figure/eval_pipeline_normal.png}
%         \caption{Normal model evaluation pipeline.}
%         \label{fig:eval_pipeline:normal}
%     \end{subfigure}
%     \\
%     \begin{subfigure}{0.47\textwidth}
%         \centering
%         \includegraphics[width=8.3cm]{figure/eval_pipeline_optimized.png}
%         \caption{Our optimized model evaluation pipeline.}
%         \label{fig:eval_pipeline:ours}
%     \end{subfigure}
%     \caption{Illustration of the normal and optimized model evaluation pipelines. We adopt layer-wise inference to make more room for pipeline parallelism, and we cache common intermediate features across different models to reduce repetitive inference. \yuanchun{This figure needs to be updated to reflect the actual pipeline.}}
%     \label{fig:eval_pipeline}
% \end{figure}

\subsection{Dynamic Model Update}
\label{sec:on-edge:update}

% \yuanchun{Extend. Should include 1. when to update (how to detect the changes); 2. how to update (using the saved candidates, how to select the next candidate, etc.); 3. when to stop updating (finishing criteria).}

The optimal subnet found by search is used in the target environment for serving. However, the subnet may become suboptimal at runtime upon environment change.

% The search process produces an optimal subnet
\name deals with environment change by dynamically paging in and paging out alternative blocks. In order to provide subnets of different latency-accuracy trade-offs, we maintain a subnet pool during searching (Section~\ref{sec:on-edge:search}) and save the $[arch, latency, accuracy]$ tuples of all the searched subnets, where $arch$ denotes the encoded architecture of the subnet. After searching, we save the subnet architectures that achieved the highest accuracy at different levels of latency (within the latency range $[T_{budget} - \Delta T, T_{budget} + \Delta T]$). For each of these subnet architectures, we save the relative latency as compared with the current optimal subnet.
% of these  instead of real latency based on the intuition that the latency of a model is roughly proportional to its number of parameters. 

At runtime, a latency monitor runs to detect the latency change of the running model. When the inference latency exceeds the budget, the latency monitor reports the latency scaling ratio $r = \frac{actual\ latency}{estimated\ latency}$, and searches in the subnet pool to find the subnet that achieves the highest accuracy within the scaled latency budget $T_{budget}/r$.
Similarly, when the actual latency is smaller than the largest relative latency in the subnet pool, the monitor also replaces the running model with a better one. 
% Note that the monitor updates the ratios in the pool after every subnet replacement.  
If the environment change is too significant and no subnet in the pool can fulfill the latency budget, we restart the search process to obtain the new optimal subnet and subnet pool.

% \subsection{Dealing with Data Heterogeneity}

% \yuanchun{Remove this subsection if we don't have time to do this. This subsection can also be merged into Section 3.2}

%% file: tex/implementation.tex
\section{Implementation}
\label{sec:implementation}

% \yuanchun{Systems paper usually includes an implementation section to describe the engineering aspects.}

% \yuanchun{A high-level introduction of the implementation.}

We implement our method using Python and Java. The on-cloud elastification part and on-device searching part are developed with PyTorch and PyTorch Mobile \cite{torch}.
% \sq{citation for pytorch}
% We also use the PyTorch Image Models \cite{rw2019timm} and EfficientDet (Pytorch) \cite{efficientdetrw} and Segmentation Models \cite{segment_github} library for pre-trained models.

\textbf{Handing two-stage models.}
% (e.g. How we deal with detection models.)}
Some deep learning applications such as object detection and semantic segmentation often require two-stage training, \eg pretraining the backbone on ImageNet \cite{imagenet} and fine-tuning on the smaller task dataset. When a DNN model needs to be trained on two datasets, \name uses a two-stage elastification strategy. Let $\mathcal{N}$ denote the DNN model well-trained on two datasets $\{D_1, D_2\}$ in order. We first elasticize the backbone of $\mathcal{N}$ and train the newly added branches on $D_1$ based on feature-based distillation (Section 4.2.1) method. After distillation, we connect the elasticized backbone to the head of $\mathcal{N}$ to make it a supernet, and further train it on $D_2$ (Section 4.2.2).   

% omit this paragraph since we should have mentioned the similar thing in supernet generation.
% \textbf{Controlling the model search space.}
% % (We limit the number of branches for large models to reduce the subnet space.)}
% Some models contain a large number of blocks. We need to limit the number of branches to reduce the subnet space and the training cost. For example, elasticizing ResNet101 into a supernet containing two levels of alternative shrunk blocks and two levels of merged blocks can create more than $2.57 \times 10^{17}$ subnets, leading to large training and searching ovehead. We observe that only including merged blocks can create enough subnets ($3.35 \times 10^8$). Merged blocks have priority over pruned blocks because they can achieve better latency-accuracy trade-offs according to our experience.

\textbf{Devices with limited memory.} The supernet generated by our method is about 2$\times$-5$\times$ larger than the pretrained model, which may not fit in the memory of some low-end devices. We use block-wise loading and inference to reduce the memory overhead. Specifically, only the blocks required by the current subnet are loaded into the memory during searching, and others are retained in the disk. Therefore, \name requires no more memory in the on-edge stage than that required by the optimal subnet.
% We need to highlight that we do not require much memory.

% \yuanchun{Add other details of several aspects that worth mention, \eg}

% \textbf{Modularized Model Loading and Evaluation.}

% \textbf{Collaboration with Pruning and Quantization.}

%% file: tex/experiment.tex
\section{Evaluation}
\label{sec:experiment}

% We implement on-cloud elasticizing and on-device searching using PyTorch and PyTorch Mobile respectively.

We conduct experiments to answer the following research questions:
(1) Is \name able to generate models with better latency-accuracy tradeoffs? (\S\ref{eval:scaling-performance}, \S\ref{eval:other-tasks})
%  \sq{uniform the usage of \S XX or Section XXX}
(2) Can \name utilize the edge data distribution? (\S\ref{eval:data-distribution})
(3) What's the efficiency of \name in both on-cloud and on-edge stages? (\S\ref{eval:training}, \S\ref{eval:on-device})

\subsection{Experimental Setup}
\label{eval:setup}

\textbf{Edge environments.} 
% (hardware and data)
We use three edge devices including an Android Smartphone (Xiaomi 12) with Snapdragon 8 Gen 1 CPU and 8 GB memory, a Jetson Nano with 4 GB memory, and an edge server with NVIDIA 3090 Ti with 24 GB GPU memory.
The batch sizes are set to 1, 1, and 32 on the three devices to simulate real workloads.
We use different latency budgets to simulate intra-device hardware diversity. The data distribution diversity is not considered in most experiments to fairly compare with the baselines. In Section~\ref{eval:data-distribution}, we use Dirichlet distribution to generate edge data, the same setting used in most Federated Learning research \cite{fed-dirichlet, yu2023multimodal, fed_dirichlet2}.
% Xiaomi 12 phone runs Android 12.0, Jetson Nano runs Jetson-inference docker and NVIDIA 3090 Ti runs Ubuntu 21.10. 

\textbf{Baselines.}
LegoDNN \cite{legodnn} and NestDNN \cite{nestdnn} are the most relevant baselines of our work. LegoDNN \cite{legodnn} is a pruning-based, block-grained technique for model scaling. NestDNN \cite{nestdnn} generates multi-capacity DNN models using filter pruning and recovering methods.  Since the source code of NestDNN is unavailable and it underperforms LegoDNN \cite{legodnn}, we conduct a detailed comparison with LegoDNN.
We also include three methods that can be used for on-device model generation, including Slimmable Networks \cite{slim,us_slim}, FlexDNN \cite{flexdnn} and SkipNet \cite{skipnet}. 
Slimmable Networks \cite{slim, us_slim} design models whose widths can be flexibly changed without retraining, FlexDNN \cite{flexdnn} is an input-adaptive method that supports early exits, and SkipNet \cite{skipnet} is a representative dynamic neural network that can dynamically switch different routes for different inputs. We adapt SkipNet \cite{skipnet} by letting it search for an optimal fixed route on the target device as the generated model. And we adapt FlexDNN \cite{flexdnn} by specifying an early exit layer for each latency budget.
We also compare with the EfficientNetV2 series \cite{efficientnetv2}, which are examples of state-of-the-art models generated by on-cloud NAS.
%To achieve end-to-end performance comparison, we directly compare the inference latency with the baselines across all experiments.

\textbf{Tasks, Models, and Datasets.}
We evaluate the performance of \name on three common vision tasks.

\begin{itemize}
\item \textbf{Image classification} aims to recognize the category of an image. We select three popular classification models, MobileNetV2 \cite{mbv2}, ResNet50 \cite{resnet}, and ResNet101 \cite{resnet} to represent small, middle, and large models. The dataset used in this task is ImageNet2012 \cite{imagenet}.
%  which contains over 1.28 million training images across 1000 image categories. We mainly compare the latency-accuracy tradeoffs with the baselines on this task.
\item \textbf{Object detection} aims to detect objects in an image, predicting the object bounding boxes and categories. We choose EfficientDet \cite{efficientdet} with ResNet50 \cite{resnet} backbone as the detection model, which is one of the top-performing detection models, and COCO2017 \cite{coco} as the dataset.
%  which contains 0.33 million images and 80 categories. 
The performance of detection models is measured by mean average precision over Intersection over Union threshold 0.5 (mAP@0.5).
\item \textbf{Semantic segmentation} aims to predict the class label of each pixel in an image. %It is widely used in medical image analysis \cite{sem-medical}, autonomous vehicle \cite{seg-ad}, and robot navigation \cite{seg-robot}.
%  semantic segmentation models are often fully convolutional networks with encoder-decoder structure. 
We choose FPN \cite{fpn} model with ResNet50 \cite{resnet} encoder pretrained on ImageNet2012 \cite{imagenet}. The dataset is CamVid \cite{camvid}, a road scene understanding dataset. The performance is measured by Mean Intersection over Union (mIoU).
\end{itemize}

% \sq{Directly evaluating on the public dataset might not align with the idea mentioned before: hence best leveraging the data from edge devices.}
% \sq{How about discussing more on how we select or s y   the evaluation dataset?}

\subsection{General Model Scaling Performance}
\label{eval:scaling-performance}

\begin{figure*}
    \centering
    \includegraphics[width=15.4cm]{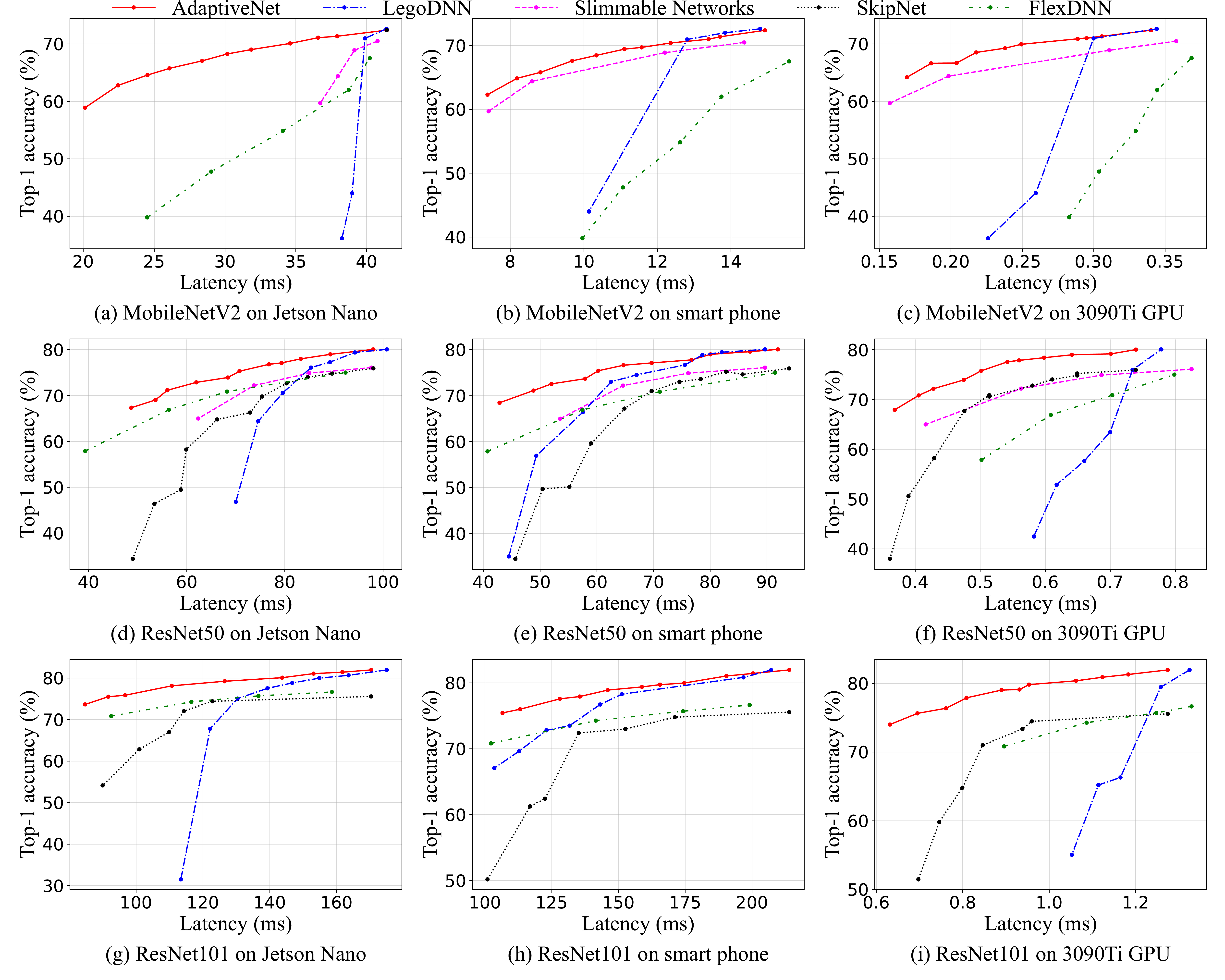}
    % \vspace{-0.3cm}
    \caption{The latency-accuracy tradeoffs of models generated by different techniques on the target devices.}
    % \vspace{-0.3cm}
    \label{fig:tradeoff-bs1}
\end{figure*}

We first evaluate the quality of models generated by our method and the baselines.
Specifically, we elasticize MobileNetV2, ResNet50, and ResNet101 which represent small, medium, and large models respectively.
For small models, we elasticize them into supernets containing five types of replaceable blocks $\{B_i^{(1)}, B_i^{(2)}, B_i^{(0)}, B_i^{(-1)}, B_i^{(-2)}\}$. The pruning rate of $B_i^{(-1)}, B_i^{(-2)}$ are 0.5 and 0.25 respectively.
For medium and large models such as ResNet50 and ResNet101, we elasticize them into supernets that only contain original and merging optional blocks $\{B_i^{(1)}, B_i^{(2)}, B_i^{(0)}\}$. After elasticizing, the supernets contain $2.58\times10^{8}, 1.06\times10^{5}, 2.57\times10^{17}$ subnets, respectively.
To make a fair comparison, we divide the validation set into two subsets, one smaller subset (3000 images) to search for the optimal subnet under 10 latency budgets, and the rest to evaluate the optimal subnet.
% Since LegoDNN \cite{legodnn} requires to build an accuracy predictor, we use the same smaller subset to build the accuracy predictor and the rest for evaluation.
% For Slimmable Networks \cite{us_slim} and SkipNet \cite{skipnet}, we use the pretrained model from their official website.

The result is displayed in Figure \ref{fig:tradeoff-bs1}, AdaptiveNet achieves higher accuracy than baseline approaches at almost every latency budget, and increases accuracy by 10.44\% and 28.03\% on average compared to LegoDNN with 90\% and 70\% latency budget respectively. This is because our elastification creates better search space of subnets and the two-stage training technique allows subnets to learn from both the original pretrained model and the labels. Thus, AdaptiveNet can outperform LegoDNN which only trains the descendent blocks to mimic the original blocks.

Besides, we observe that AdaptiveNet outperforms the baseline models more at a lower latency budget. At the 60\% and 80\% latency budget, AdaptiveNet achieves 42.53\% and 29.16\% higher accuracy on average respectively. This is because our approach includes merging two or more blocks into one replacement block compared to pruning-based model scaling techniques. Such block merging can save more latency with a smaller loss of accuracy than high-ratio pruning. 

%Compared to SkipNet \cite{skipnet} and FlexDNN \cite{flexdnn}, our method achieves better accuracy-latency tradeoffs for two reasons. First, block-skipping and early-exiting methods only gain benefits by saving computation by reducing the depth of DNN models. \name, however, can reduce both the depth and width of a DNN model by introducing merged and pruned blocks. Second, input-adaptive methods are not completely suitable for situations where latency budgets are strict. In order to meet a lower latency budget, input-adaptive methods have to skip more blocks, which have been trained specifically to deal with difficult inputs.

We also notice that the gap between \name and Slimmable Networks is small on smartphones and 3090 Ti. The main reason is that slimmed networks can better utilize the computational resources on such devices. However, because the slimmable models are based on custom backbones, they cannot support SOTA pretrained models and are not flexible for normal developers to use.

Further, \name can be used with multiple pretrained models to achieve more wide-range and fine-grained trade-offs. 
Figure~\ref{fig:nas}(a) shows the performance of models generated from two oracle EfficientNetV2 models, where \name provides over 20 meaningful latency-accuracy trade-offs between the oracle models.
Thus, developers can use \name as an effective supplement to manually-created or cloud-generated models to offer more choices for the edge with little overhead (dozens of hours).

\subsection{Performance on Other Tasks}
\label{eval:other-tasks}

We also test AdaptiveNet on object detection and semantic segmentation to evaluate its generalizability and performance on complex two-stage tasks (pretrained on ImageNet2012 \cite{imagenet}, fine-tuned on COCO2017 \cite{coco} and CamVid \cite{camvid}). Our object detection model, EfficientDet \cite{efficientdet}, consists of a backbone, neck, and head, among which the backbone takes up most of the inference latency (more than 90\% according to our measurements), thus we only elasticize the backbones. For the same reason, we only elasticize the encoder of FPN \cite{fpn}.
Since the official code of our baseline LegoDNN \cite{legodnn} on object detection and semantic segmentation cannot run properly, we implement the training process of LegoDNN \cite{legodnn} on both tasks.
To make fair comparisons, \name and LegoDNN \cite{legodnn} start from the same pretrained model and train for the same GPU hours. After training, we randomly sample the same number (500) of subnets for both tasks and evaluate them on the test set.

\begin{figure}
    \centering
    \includegraphics[width=8cm]{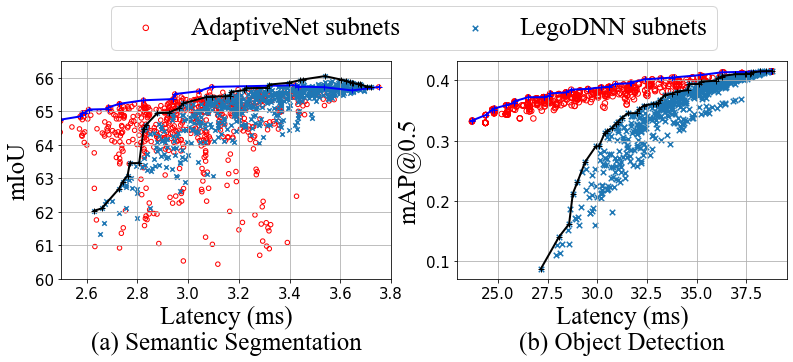}
    \vspace{-0.3cm}
    \caption{Quality of models generated for detection and segmentation tasks.}
    \vspace{-0.1cm}
    \label{fig:othertasks}
\end{figure}

The results are shown in Figure \ref{fig:othertasks}. Similar to the classification tasks, \name achieves reasonable scaling performance and outperforms the baseline. Some of the FPN subnets can even achieve better tradeoffs than the original pretrained model, which is because the original model is over-fitted. Our subnets generated by merging some original blocks together can reduce the parameter size of the original model, which reduces over-fitting and improves accuracy.
% \name outperforms LegoDNN \cite{legodnn} on both tasks with \xx and \xx higher performance respectively.

% \begin{figure}
%     \begin{subfigure}[b]{0.24\textwidth}
%         \centering
%         \includegraphics[width=4.4cm]{figure/detection.PNG}
%         \caption{Object Detection}
%         \label{fig:strategy}
%     \end{subfigure}
%     ~~
%     \centering
%     \begin{subfigure}[b]{0.24\textwidth}
%         \centering
%         \includegraphics[width=4.4cm]{figure/segmentation.PNG}
%         \caption{Semantic Segmentation}
%         \label{fig:end2end}
%     \end{subfigure}
%     \caption{Quality of models generated for detection and segmentation tasks.}
%     \label{fig:othertasks}
% \end{figure}

\subsection{Impact of Edge Data Distribution Shift}
\label{eval:data-distribution}

\begin{figure*}
    \centering
    % \begin{subfigure}[b]{0.32\textwidth}
    %     \centering
    %     \includegraphics[width=6.5cm]{figure/iid.png}
    %     \caption{IID data distribution.}
    %     \label{fig:iid}
    % \end{subfigure}
    % ~~
    % \begin{subfigure}[b]{0.32\textwidth}
    %     \centering
    %     \includegraphics[width=6.5cm]{figure/alpha001.PNG}
    %     \caption{Dirichlet data distribution ($\alpha=0.01$).}
    %     \label{fig:alpha001}
    % \end{subfigure}
    % ~~
    % \begin{subfigure}[b]{0.32\textwidth}
    %     \centering
    %     \includegraphics[width=6.5cm]{figure/alpha01.PNG}
    %     \caption{Dirichlet data distribution ($\alpha=0.1$).}
    %     \label{fig:alpha01}
    % \end{subfigure}
    \includegraphics[width=16.5cm]{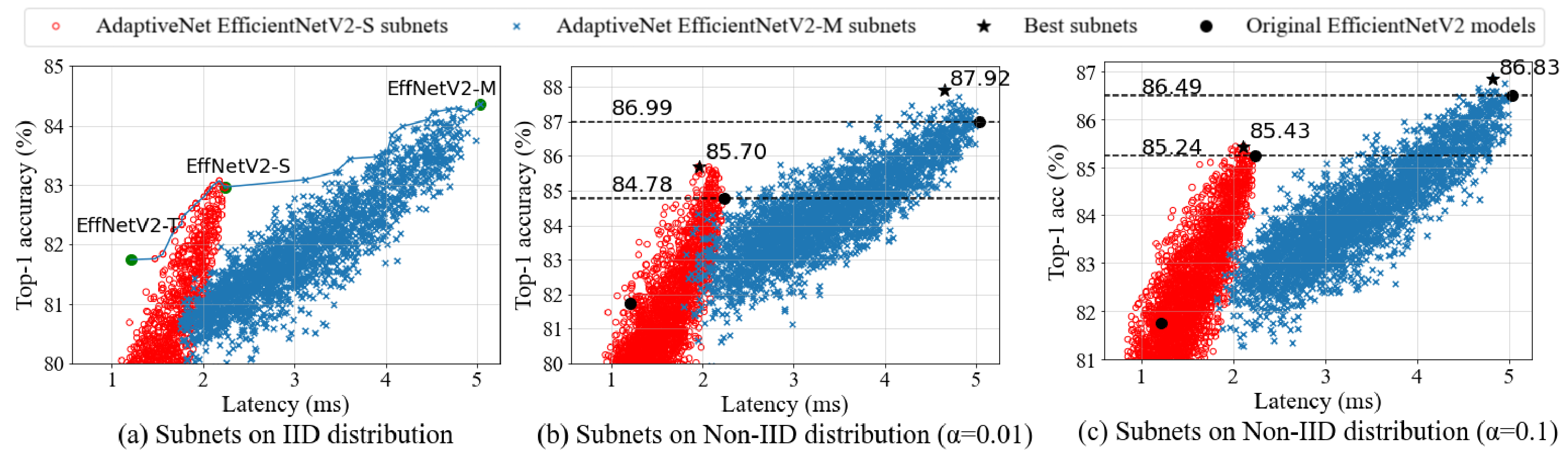}
    \vspace{-0.3cm}
    \caption{Quality of models generated by \name on different edge data distribution in comparison with cloud-generated oracle models (Latency measured on NVIDIA 3090 Ti GPU).}
    \vspace{-0.3cm}
    \label{fig:nas}
\end{figure*}

Since \name generates the model directly on the target device, it can utilize edge data as compared with on-cloud NAS. 
We examine the quality of models generated by \name on different edge datasets simulated with Dirichlet distributions.
The results are shown in Figure \ref{fig:nas}.
% \name is an effective supplement to traditional NAS in two ways. 1) \name can elasticize the searched optimal model and achieve more fine-grained accuracy-latency trade-offs with little overhead (dozens of GPU hours) compared to extensive NAS overhead (thousands of GPU hours \cite{ofa}). The elastification result is shown in Figure \ref{fig:iid}, \name provides over 20 latency-accuracy trade-offs between EfficientNetV2-M, EfficientNetV2-S and EfficientNetV2-Tiny;

We notice that \name can outperform the EfficientNetV2 models that are generated by extensive on-cloud NAS \cite{efficientnetv2} on unbalanced edge data distributions. 
Some of the subnets may improve the top-1 accuracy by 1.07\% while saving 7.71\% latency at the same time, which is a hard-won improvement since the original model has achieved excellent performance (with 86.99\% top-1 accuracy and 5.03ms latency). We also observe that the advantage of \name increases when the data distribution is more unbalanced. 

Given the fact that the edge data distributions are usually different from training ones \cite{personalizedfedlearning}, we believe the post-deployment model generation mechanism of \name is a more promising direction to seek in edge AI scenarios.

% \vspace{-0.1cm}
\subsection{On-cloud Training Performance}
\label{eval:training}

\begin{figure}
    \centering
    \includegraphics[width=5.8cm]{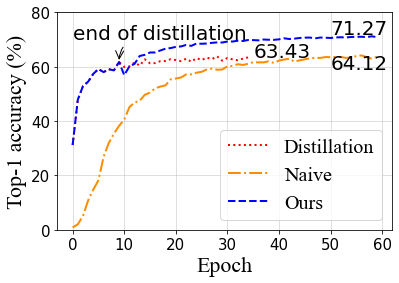}
    \vspace{-0.4cm}
    \caption{Training efficiency of on-cloud elastification.}
    \label{fig:distillation_gain}
    \vspace{-0.1cm}
\end{figure}

We examine the efficiency and effectiveness of our on-cloud elastification stage.
We compare our supernet training method with a supervised training baseline (\ie our method without distillation) and a distillation-only baseline (\ie our model without whole-model tuning) and plot the progressive top-1 accuracy on ImageNet in Figure \ref{fig:distillation_gain}.
Although all of the training methods can converge after 50 epochs, our two-step training technique is 7.15\% and 7.84\% higher than using supervised training only and distillation only.
% The naive approach indicates uniformly sampling subnets with random initialization. Weight transfer method indicates initializing optional blocks with the first blocks they substitute. 
% This method can be viewed as a block-wise pruning and retraining, so it outperforms simply training from the scratch. 
Our supernet training also converges faster than the baselines with only 60 epochs (about 13 hours), which is also significantly faster than on-cloud NAS (>1200 GPU hours \cite{ofa}).

% \vspace{-0.1cm}
\subsection{On-device Adaptation Efficiency}
\label{eval:on-device}

% \begin{table}
% 	\caption{On-device Search Overhead Comparison}
% 	\centering
% 	\begin{tabular}{llll}
% 		\toprule
% 		%\multicolumn{2}{c}{Part}                   \\
% 		\cmidrule(r){1-2}
% 		Device     & Model     & Search Technique  & Overhead (Hour)\\
% 		\midrule
% 		Nano                  & MobileNetv2  & GA    & 2.477\\
% 		Nano                  & MobileNetv2  & Ours    & 0.671 \quad(-72.9\%)\\
% 		Nano                  & ResNet50     & GA      & 4.734 \\
% 		Nano                  & ResNet50     & Ours    & 2.390\quad(-49.5\%) \\
% 		2080Ti                & MobileNetv2  & GA   &  \\
% 		2080Ti                & MobileNetv2  & Ours    & \\
% 		2080Ti                & ResNet50     & GA    & \\
% 		2080Ti                & ResNet50     & Ours    & \\
% 		\bottomrule
% 	\end{tabular}
% 	\label{tab:search}
% \end{table}

\begin{figure}
    \centering
    \includegraphics[width=7cm]{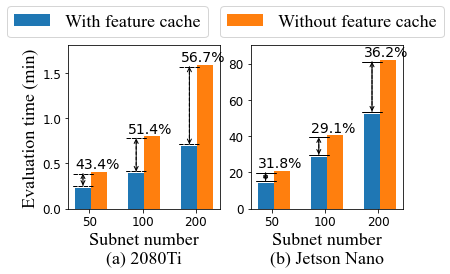}
    \vspace{-0.3cm}
    \caption{Speed of evaluating a group of subnets.}
    \vspace{-0.1cm}
    \label{fig:accelerate}
\end{figure}

% on two devices: an NVIDIA Jetson Nano and 2080Ti representing embedded devices and edge server respectively. We select elasticized ResNet50 as supernet to be searched. 

\begin{figure}
    \begin{subfigure}[b]{0.24\textwidth}
        \centering
        \includegraphics[width=4.4cm]{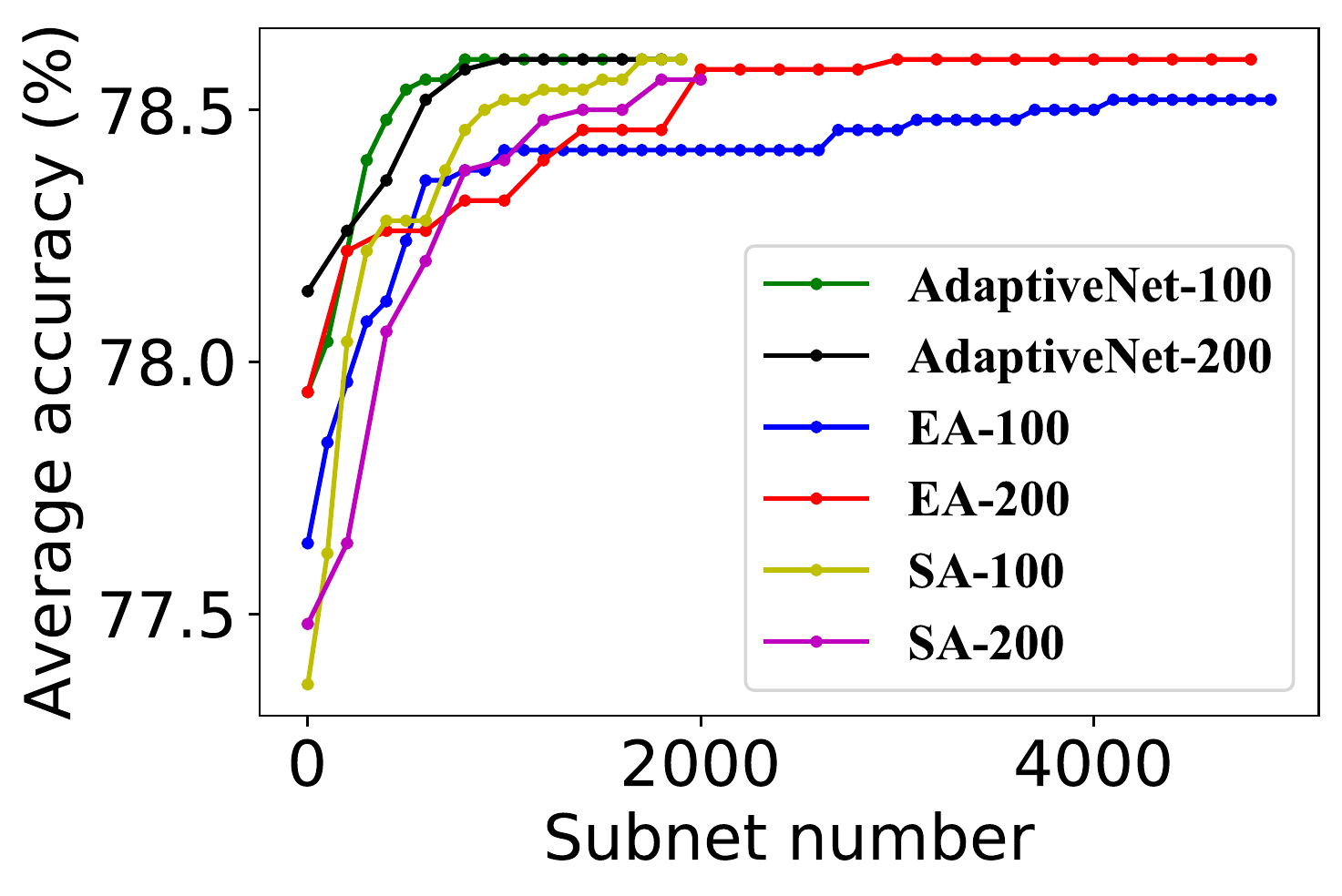}
        \vspace{-0.3cm}
        \caption{Optimal accuracy achieved with different num of subnets.}
        % \vspace{-0.3cm}
        \label{fig:strategy}
    \end{subfigure}
    ~~
    \centering
    \begin{subfigure}[b]{0.24\textwidth}
        \centering
        \includegraphics[width=4.4cm]{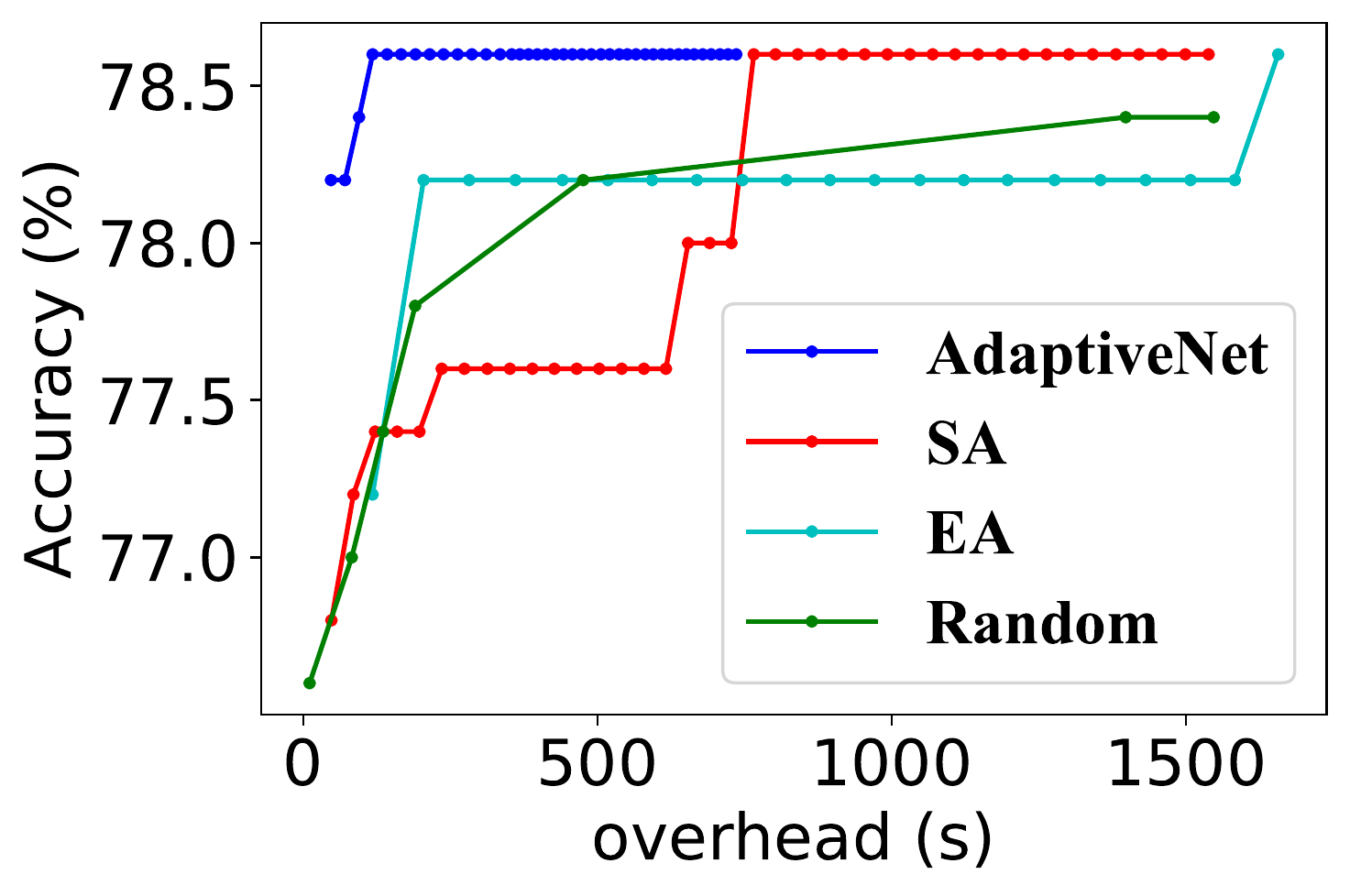}
        % \vspace{-0.3cm}
        \caption{Optimal accuracy achieved with different search time.}
        % \vspace{-0.3cm}
        \label{fig:end2end}
    \end{subfigure}
    \vspace{-0.5cm}
    \caption{Comparison of search efficiency between different methods.}
    \label{fig:eval:search-efficiency}
\end{figure}

% There are three main classes of search methods in NAS, including reinforcement learning \cite{nas-rl, nasrl}, evolutionary algorithm \cite{single_path_one_shot, ofa} and gradient-based methods \cite{darts-gradient-based, proxylessnas}.
% On-device training is usually a difficult job for edge or mobile devices, so we choose evolutionary algorithm as our searching method. 
In this section, we evaluate the performance of the on-device search in \name.
%There are multiple ways to do subnet search in conventional NAS, but 
Most of the subnet search methods in conventional NAS are too heavy for edge devices (\eg reinforcement learning \cite{nas-rl, nasrl} and gradient-based methods \cite{darts-gradient-based, proxylessnas}).
So we choose normal evolutionary search \cite{evolutionarysearch} and simulated annealing \cite{foxnas} as our baselines.
To examine the effectiveness of our method, we conduct two ablation experiments and one end-to-end experiment. All the experiments use the same 500 images for searching.

Figure \ref{fig:accelerate} shows the acceleration percentage of our reuse-based model evaluation method (Section~\ref{sec:on-edge:evaluate}). We compare the time spent to evaluate 50, 100, and 200 subnets respectively. 
% To make fair comparisons, we use layer-wise inference strategy for both methods. 
Our method saves up to 56.7\% and 36.2\% of search overhead compared to normal evaluation pipeline, and consumes 100-500MB of memory. It is notable that the memory cost of our method is controllable by adjusting the depth of the subnet tree. If there is no space for feature maps, we can set the depth to 0, which will reduce the GPU memory overhead to zero with some sacrifice of search efficiency.

Figure \ref{fig:strategy} shows the benefits of our model-guided search strategy. To achieve the same average accuracy, \name, evolutionary algorithm, and simulated annealing need to try 800, 3100, and 1800 subnets respectively.
% indicating that our method improves the efficiency by 74.2\% and 55.6\%.
Figure \ref{fig:end2end} shows the end-to-end search efficiency comparison between \name and baselines on NVIDIA 2080 Ti. We conduct three individual experiments with a population size of 50, 100, and 200, respectively, and show the best results for each strategy. \name, simulated annealing algorithm, and evolutionary algorithm to find optimal subnet in 117.6, 765.6, and 1656.9 seconds respectively, indicating that our method can improve search efficiency by more than 80\%.

\begin{table}
	\caption{Size (MB) of AdaptiveNet-generated supernets and their corresponding pretrained models.}
	% \vspace{-0.3cm}
	\centering
	\resizebox{.35\textwidth}{!}
    {
	% \begin{tabular}{crrr}
	% 	\toprule
	% 	%\multicolumn{2}{c}{Part}                   \\
	% 	\cmidrule(r){1-2}
	% 	Model    &  MobileNetV2   & ResNet50 & ResNet101 \\
	% 	\midrule
	% 	\name & 32.88 & 381.66 & 503.82 \\
	% 	Pretrained Model & 14.20 & 102.48 & 178.71 \\

	% 	\bottomrule
	% \end{tabular}
        \begin{tabular}{crr}
		\toprule
		%\multicolumn{2}{c}{Part}                   \\
		\cmidrule(r){1-2}
		Model    &  Pretrained model & Supernet  \\
		\midrule
		MobileNetV2 & 14.20   & 32.88  \\
		ResNet50    & 102.48  & 381.66\\
        ResNet101   & 178.71  & 503.82 \\
		\bottomrule
	\end{tabular}
	}
 \vspace{-0.4cm}
	\label{tab:size}
\end{table}   

\textbf{Network transmission overhead.} Table \ref{tab:size} shows the size of \name and pretrained models. Although \name increases the size of models, we believe it can actually save network overhead. \name only needs to transmit the supernet to edge devices once, which is 1.32$\times$-2.72$\times$ larger than the original pre-trained model. However, to achieve similar performance, conventional model deployment approaches have to collect device information and re-transmit the model when the edge environment changes, which is $n\times$ larger than the original model, where $n$ is the time of changes.% On-device search cost is another key overhead. The latency of subnet searching is evaluated above. The memory overhead is controllable by tuning the depth of the “subnet tree”, which determines the number of feature maps cached in memory. If there is no space for feature maps, we can set the depth to 0. In our experiments, we set the depth to 2$\sim$4, which only consumes 300$\sim$500MB of memory. }

\begin{figure}
    \centering
    \includegraphics[width=7cm]{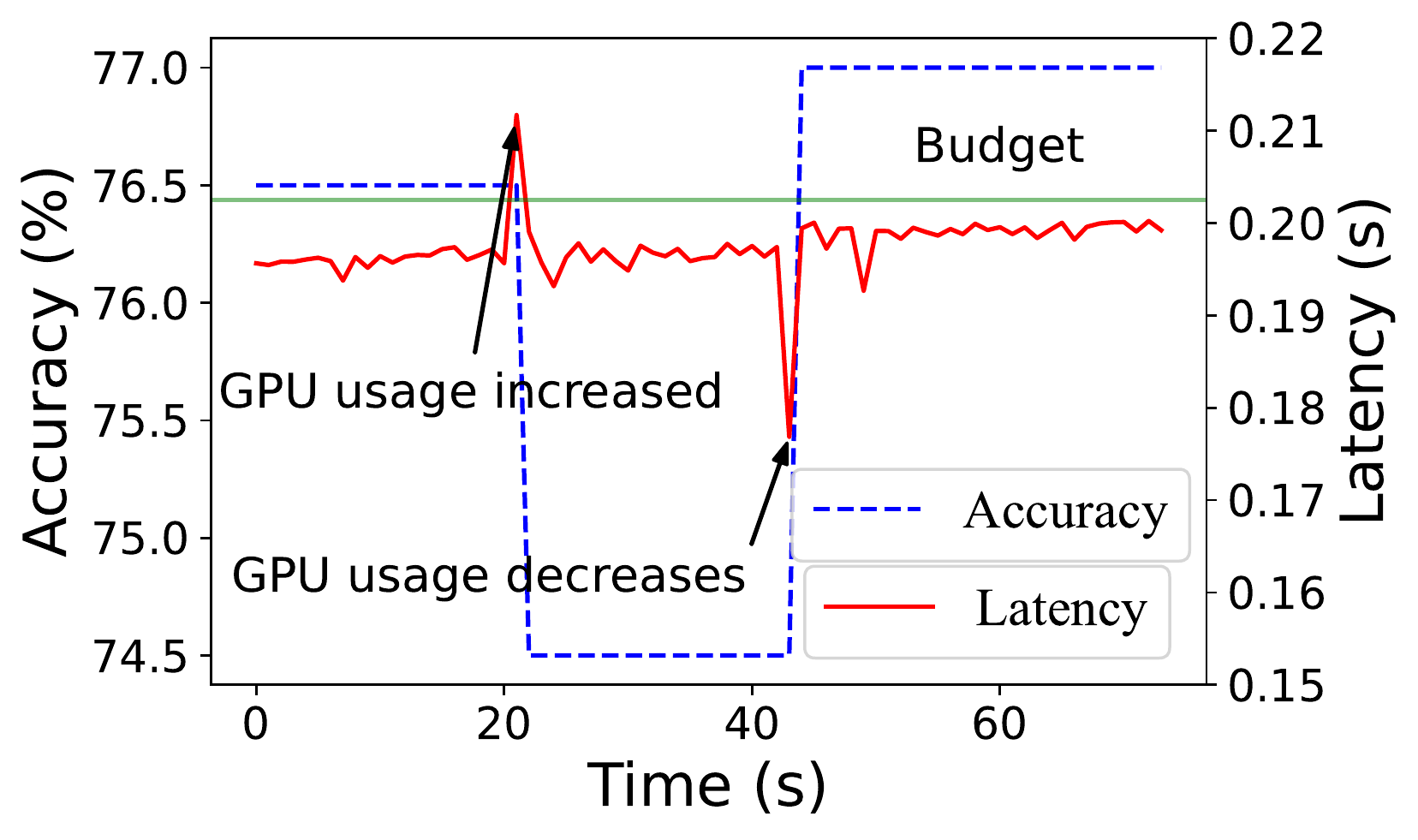}
    % \vspace{-0.5cm}
    \caption{Demonstration of dynamic model update.}
    % \vspace{-0.1cm}
    \label{fig:monitor}
\end{figure}

\textbf{Real-time Model Update Efficiency.}
We further test the performance of our dynamic model update module and present the result in Figure \ref{fig:monitor}. We choose ResNet50 \cite{resnet} as our pretrained model and the experiment is conducted on NVIDIA 2080 Ti. We adjust the GPU usage by running and killing MobileNetV2 inference processes.
Our dynamic model update is fast and responsive. After the latency budget is exceeded or under-utilized, \name can find the optimal model and recover the latency within 1 second. Specifically, we obtain a pool of optimal subnets for a range of latency budgets during the on-device search. The runtime update module only needs to switch to the proper model, instead of searching from scratch. Thus, it should be easy for our method to catch up with the workload dynamics. Besides, if the real workload is very dynamic, we can control the update frequency to avoid overreaction.

%% file: tex/conclusion.tex
\section{Discussion}

An issue that may be a concern in the on-device model generation is the need for labeled edge data, which might be difficult if the data is not auto-labeled (like in many unsupervised tasks \cite{yu-etal-2018-device,unsupervised1}).
Such a dataset can be generated by querying an oracle model with unlabelled edge data. Letting the cloud send public data to the edge is also an option, although the edge data characteristics will not be utilized in this way.

% \wh{Although we mainly evaluate the performance of \name on vision tasks, this method should also generalize to other tasks such as NLP. Transformer models \cite{transformer} are also composed of repeated blocks such as encoders and decoders, so we should be able to elasticize it into a supernet and choose the optimal subnet from it at edge devices. The high redundancy of model parameters in NLP models may create a larger space for elastification.
% }
Although AdaptiveNet is mainly evaluated on vision tasks, it should be able to generalize to other tasks such as NLP. Transformer models \cite{transformer} are also composed of repeated blocks such as encoders and decoders, so we should be able to elasticize them into supernets and choose the optimal subnet from them at edge devices.

We also want to discuss the relationship between AdaptiveNet and on-device training, which can be used to improve model quality after deployment. First, on-device training typically requires heavy computation and sufficient training data to be effective, which \name does not require. Besides, on devices with good training conditions, a better architecture found by \name can also be beneficial.

% When the edge datasets are not big enough, on-device fine-tuning may lead to overfitting. AdaptiveNet, on the other hand, need not finetune after deployment, and a small dataset can be enough for searching for the optimal subnet. Besides, we can find a better start point for finetuning under the latency budget when the edge data distribution is different from the training data.

\section{Conclusion}

This paper proposes a novel approach for on-device, post-deployment, and environment-aware model architecture generation.
The approach is implemented as an end-to-end system equipped with on-cloud model elastification and on-device model adaptation techniques.
Experiments have demonstrated the remarkable model quality and model generation efficiency of our method.
Developers can scale their AI applications to diverse and dynamic edge environments with our system by simply specifying a pretrained model.

\section*{Acknowledgement}
\label{sec:acknowledgement}

This work is supported by the National Key R\&D Program of China (No.2022YFF0604501) and NSFC (No.62272261).